\documentclass{article}

\usepackage[main, final]{neurips_2026}

\usepackage[utf8]{inputenc} 
\usepackage[T1]{fontenc}    
\usepackage{hyperref}
\hypersetup{colorlinks=true,citecolor=brown,urlcolor=gray,linkcolor=BlueViolet}
\usepackage{url}            
\usepackage{booktabs}       
\usepackage{amsfonts}       
\usepackage{nicefrac}       
\usepackage{microtype}      
\usepackage{xcolor}
\usepackage[dvipsnames]{xcolor}
\usepackage[table]{xcolor}
\usepackage{xspace}
\usepackage{enumitem}
\usepackage{amssymb}
\usepackage{multirow}
\usepackage{makecell}
\usepackage{float}
\usepackage{listings}
\usepackage[many]{tcolorbox}
\tcbuselibrary{skins, listings, breakable}
\usepackage{xcolor}
\usepackage{wrapfig}
\usepackage{tabularx}
\usepackage{fontawesome5}

\usepackage[normalem]{ulem}

\definecolor{promptbg}{HTML}{F8F9FA}
\definecolor{promptframe}{HTML}{343A40}

\definecolor{trainedbg}{HTML}{F0FDF4}
\definecolor{trainedframe}{HTML}{15803D}

\definecolor{basebg}{HTML}{FEF2F2}
\definecolor{baseframe}{HTML}{B91C1C}

\definecolor{xmltag}{HTML}{005CC5}       
\definecolor{xmlstring}{HTML}{032F62}    
\definecolor{pykeyword}{HTML}{D73A49}    
\definecolor{codecomment}{HTML}{6A737D}  
\lstdefinelanguage{XMLCustom}{
    basicstyle=\ttfamily\scriptsize\color{black},
    moredelim=[s][\color{xmltag}\bfseries]{<}{>}, 
    morestring=[b]",
    stringstyle=\color{xmlstring},
    identifierstyle=\color{black},
    breaklines=true,
    breakatwhitespace=true,
    showspaces=false, 
    showstringspaces=false,
    columns=fullflexible,
    keepspaces=true
}
\lstdefinelanguage{BaseHybrid}{
    basicstyle=\ttfamily\scriptsize\color{black},
    moredelim=[s][\color{xmltag}\bfseries]{<}{>},
    keywordstyle=\color{pykeyword}\bfseries,
    morestring=[b]",
    morestring=[b]',
    stringstyle=\color{xmlstring},
    morecomment=[l]{\#},
    commentstyle=\color{codecomment}\itshape,
    breaklines=true,
    breakatwhitespace=true,
    showspaces=false,
    showstringspaces=false,
    columns=fullflexible,
    keepspaces=true
}
\newtcolorbox{promptbox}[1][]{
    enhanced, breakable, colback=promptbg, colframe=promptframe, boxrule=0.8pt,
    arc=3mm, left=4mm, right=4mm, top=3mm, bottom=3mm,
    title= \textbf{#1}, coltitle=white, fonttitle=\sffamily,
    attach boxed title to top left={xshift=5mm, yshift=-2mm}
}

\newtcblisting{trainedbox}[1][]{
    enhanced, breakable, colback=trainedbg, colframe=trainedframe, boxrule=0.8pt,
    arc=3mm, left=4mm, right=4mm, top=3mm, bottom=3mm,
    title= \textbf{#1}, coltitle=white, fonttitle=\sffamily,
    attach boxed title to top left={xshift=5mm, yshift=-2mm},
    listing only, listing options={language=XMLCustom}
}

\newtcblisting{basebox}[1][]{
    enhanced, breakable, colback=basebg, colframe=baseframe, boxrule=0.8pt,
    arc=3mm, left=4mm, right=4mm, top=3mm, bottom=3mm,
    title= \textbf{#1}, coltitle=white, fonttitle=\sffamily,
    attach boxed title to top left={xshift=5mm, yshift=-2mm},
    listing only, listing options={language=BaseHybrid}
}

\newcommand{\policy}{\ensuremath{\pi_{\vtheta}}\xspace}
\newcommand{\traj}{\ensuremath{\vtau}\xspace}
\newcommand{\rmModel}{\ensuremath{r_{\vphi}}\xspace}
\newcommand{\chosentraj}{\ensuremath{\vtau^{+}}\xspace}
\newcommand{\rejectedtraj}{\ensuremath{\vtau^{-}}\xspace}
\newcommand{\basepolicy}{\ensuremath{\pi_{0}}\xspace}

\newcommand{\ours}{\texttt{Orch-RM}\xspace}

\newcommand{\update}[1]{%
  \textcolor{blue}{\sf{[Update: #1]}}%
}

\usepackage[normalem]{ulem}

\renewcommand{\paragraph}[1]{%
  \par\noindent\textbf{#1}\quad
}

\usepackage{amsmath,amsfonts,bm,eqnarray}
\usepackage{cancel}
\usepackage{amsthm}
\usepackage{tikz}
\usetikzlibrary{tikzmark}

\newtheorem{innercustomgeneric}{\customgenericname}
\providecommand{\customgenericname}{}
\newcommand{\newcustomtheorem}[2]{%
  \newenvironment{#1}[1]
  {%
   \renewcommand\customgenericname{#2}%
   \renewcommand\theinnercustomgeneric{##1}%
   \innercustomgeneric
  }
  {\endinnercustomgeneric}
}

\newcustomtheorem{customThm}{Theorem}
\newcustomtheorem{customLemma}{Lemma}
\newcustomtheorem{customCor}{Corollary}
\newcustomtheorem{customProposition}{Proposition}

\def\Tabref#1{Table~\ref{#1}}

\def\Figref#1{Fig.~\ref{#1}}

\def\appref#1{Appendix~\ref{#1}}
\def\Secref#1{Sec.~\ref{#1}}

\def\eqref#1{equation~\ref{#1}}

\def\1{\bm{1}}

\def\rvx{{\mathbf{x}}}

\def\vtheta{{\bm{\theta}}}

\def\va{{\bm{a}}}

\def\vx{{\bm{x}}}
\def\vy{{\bm{y}}}
\def\vz{{\bm{z}}}

\def\vphi{{\boldsymbol{\phi}}}

\def\vtau{{\boldsymbol{\tau}}}

\def\mX{{\bm{X}}}
\def\mY{{\bm{Y}}}
\def\mZ{{\bm{Z}}}

\DeclareMathAlphabet{\mathsfit}{\encodingdefault}{\sfdefault}{m}{sl}
\SetMathAlphabet{\mathsfit}{bold}{\encodingdefault}{\sfdefault}{bx}{n}

\def\gD{{\mathcal{D}}}

\def\gL{{\mathcal{L}}}

\newcommand{\E}{\mathbb{E}}

\renewcommand{\hat}{\widehat}
\renewcommand{\frac}{\tfrac}

\definecolor{lightergray}{HTML}{e5e5e5}
\definecolor{lightgray}{HTML}{cccccc}

\title{\ours: Reward Modeling for \\Multi-Agent Orchestration}

\author{%
  King Yeung Tsang\thanks{Equal contribution. 
  $^{\dagger}$Project co-lead.
  $^{\ddagger}$Equal advising.
  $^1$Rutgers University.
  $^2$Salesforce AI Research.
  $^3$University of Illinois at Urbana-Champaign.
  $^4$Nanyang Technological University.
  Correspondence to: King Yeung Tsang <kingyeung.tsang@gmail.com>, 
  Haizhou Shi <haizhou.shi@rutgers.edu>, 
  Zixuan Ke <zixuan.ke@salesforce.com>,
  Shafiq Joty <sjoty@salesforce.com>,
  Hao Wang <hwangml@illinois.edu>.}\space\space{$^{1}$}
  \\
  \And
  Zihao Zhao$^{*1,3}$\\
  \And
  Vishal Venkataramani$^{1}$\\
  \And
  Haizhou Shi$^{\dagger 1,2}$\\
  \And
  Zixuan Ke$^{\dagger 2}$\\
  \And
  Semih Yavuz$^{2}$\\
  \And
  Shafiq Joty$^{\ddagger 2,4}$\\
  \And
  Hao Wang$^{\ddagger 1,3}$\\
}

\begin{document}

\maketitle

\vspace{-0.8em}
{\centering
  \href{https://github.com/Wang-ML-Lab/OrchRM}{\faGithub\ \texttt{Code}}\hspace{2em}
  \href{https://huggingface.co/datasets/tsangkingyeung/OrchRM-datasets}{\faDatabase\ \texttt{Dataset}}\par}
\vspace{0.4em}

\begin{abstract}
Multi-Agent Systems (MAS) built on Large Language Models (LLMs) require effective orchestration to coordinate specialized agents, yet training such orchestrators is hindered by limited supervision and high computational cost. 
We propose \textbf{Orch}estration \textbf{R}eward \textbf{M}odeling (\textbf{\ours}), a framework for evaluating orchestration quality without new human annotations.
\ours leverages intermediate artifacts from multi-agent executions to construct win-lose pairs for Bradley-Terry reward model training. 
Once trained, \ours operates directly at the orchestration level, enabling efficient and high-performing reward-guided orchestrator training and MAS test-time scaling. 
As shown in \Figref{fig:fig1}, \ours improves training efficiency by up to 10$\times$ in token usage while improving MAS test-time scaling performance by up to 8\% in accuracy. 
Improvements vary across benchmarks, with larger gains on AIME and BrowseComp+, while remaining competitive on HotpotQA and GPQA. The smaller gains on the latter two benchmarks may partly reflect their limited structural diversity in orchestration plans. 
These results highlight orchestration-level reward modeling as a promising direction for scalable and robust multi-agent orchestration.
\end{abstract}


\begin{figure}[H]
    \centering
    \vspace{-0.5em}
    \includegraphics[width=0.9\linewidth]{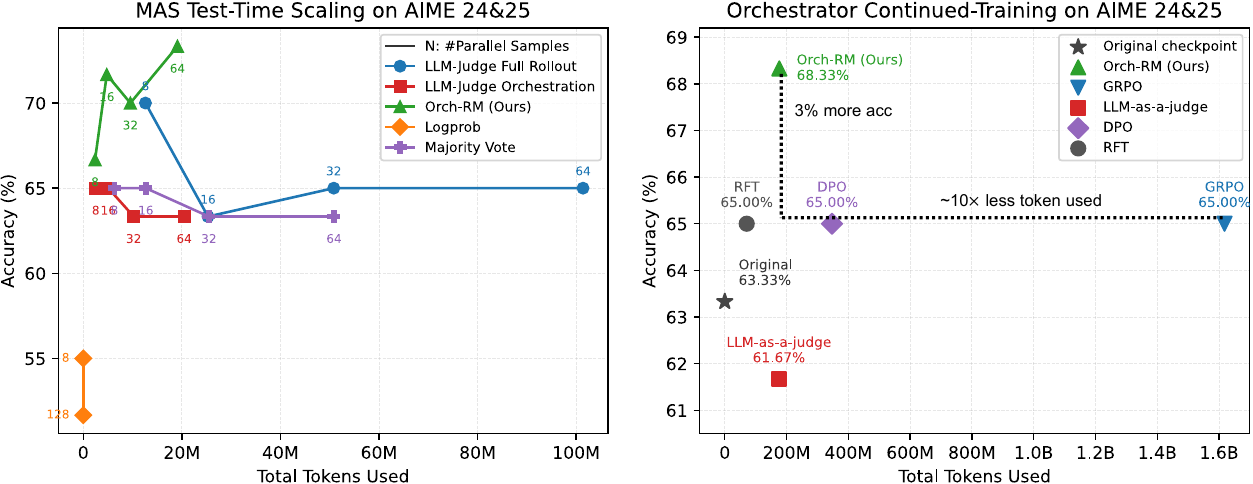}
    \vspace{-0.5em}
    \caption{
    \textbf{\ours for MAS Test-Time Scaling~(Left) and Continued Orchestrator Training~(Right).}
    \textbf{Left:} By operating directly at the orchestration level of MAS, \ours achieves a stronger accuracy-efficiency trade-off for test-time scaling, reducing token usage while improving performance over baselines. 
    \textbf{Right:} Using \ours as the reward signal for continued MAS orchestrator training yields consistent performance gains with substantially lower token usage.
    }
    \label{fig:fig1}
\end{figure}

\section{Introduction}



Multi-Agent Systems~(MAS) built on Large Language Models~(LLMs) have demonstrated strong potential for tackling complex tasks through the coordination of specialized agents~\citep{ke2025surveyfrontiersllmreasoning}. By leveraging mechanisms such as task decomposition, evidence gathering, critique, and aggregation, these systems can produce outputs that are often more robust than those of a single agent. More recent advances have shifted from manually designed multi-agent workflows to automated \emph{orchestration}, where an LLM-based controller dynamically plans the workflow structure and agent interactions required for a given task~\cite{ke2026masorchestraunderstandingimprovingmultiagent, niu2025flowmodularizedagenticworkflow,yue2026statictemplatesdynamicruntime, zhang2025aflow}. This transition enhances both the flexibility and generalizability of MAS, enabling them to adapt their planning strategies across diverse domains, including previously unseen scenarios.

As naive off-the-shelf orchestrators often underperform in specialized domains~\cite{ke2026masorchestraunderstandingimprovingmultiagent,niu2025flowmodularizedagenticworkflow, yue2026statictemplatesdynamicruntime}, developing a capable LLM-based orchestrator has become increasingly important, yet remains highly non-trivial. This difficulty stems from several factors. First, the strong interdependencies among orchestration strategies, task types, and sub-agent capabilities make it challenging to collect high-quality human annotations for training. Second, the complexity of multi-agent workflows, coupled with the accumulated variance across full sub-agent rollouts, leads to substantial computational and time costs during training; for example, the recent MAS-Orchestra framework employs GRPO training, consuming over 1B tokens for only 100 update steps of the orchestrator~\citep{ke2026masorchestraunderstandingimprovingmultiagent}.

To address the scarcity of high-quality orchestration annotations and the inefficiency of training such models, we propose \textbf{Orch}estration \textbf{R}eward \textbf{M}odeling~(\textbf{\ours}) for MAS. Rather than introducing a new orchestrator architecture or reinforcement learning objective, our goal is to investigate \emph{the feasibility of leveraging reward models} to improve both training and test-time performance in \emph{orchestration-based MAS systems}.
In contrast to Reinforcement Learning via Human Feedback~(RLHF)~\citep{ouyang2022training}, which relies heavily on large-scale human preference annotations, 
{\ours learns from automatically constructed preference pairs} to evaluate orchestration quality. Specifically, we collect intermediate artifacts, i.e., model checkpoints and generated trajectories,
from MAS-Orchestra training and, through a multi-faceted grouping strategy, construct diverse win-lose pairs of orchestration trajectories to train a Bradley-Terry reward model~\citep{zhang2026bradleyterry}. 
The resulting reward model’s ability to assess multi-agent orchestrations complements the known limitations of existing RMs and LLM-as-a-Judge methods when applied to MAS, both at the trajectory level~\cite{venkataramani2026masproveunderstandingprocessverification} and at the orchestration level (\Secref{sec:experiment-tts} and \ref{sec:experiment-training}).
We evaluate our approach on mathematical reasoning, web-based question answering, multi-hop reasoning, and out-of-distribution scientific reasoning. We integrate our learned \ours into both \emph{(i) inference through best-of-$N$~(BoN) selection for efficient test-time scaling} 
and \emph{(ii) reward-guided orchestrator training}. 
On the inference side, \ours selects orchestration plans through best-of-$N$ (BoN) scoring before sub-agent execution. Compared with majority voting, \ours achieves clear gains on most evaluated benchmarks.
On the training side, reward-guided continued training generally improves MAS-Orchestra ~\citep{ke2026masorchestraunderstandingimprovingmultiagent} over the original checkpoint across the four evaluated benchmarks.
Even training an orchestrator from scratch with reward-model feedback can approach the performance of the original teacher model. 
Together, these results suggest that our proposed orchestration reward modeling is a promising direction for both improving test-time selection and providing reusable supervision for automated multi-agent orchestration.
Our main contributions are:
\begin{itemize}[leftmargin=24pt]
    \item We developed \ours, a reward modeling framework for training LLM-based orchestrators.
    \item We integrated \ours into orchestrator training, enabling reward-guided continued optimization that improves existing models and supports training from scratch to approach the performance of RL-trained counterparts.
    \item We applied \ours at inference time, where it selects higher-quality rollouts than majority voting and enables more effective test-time scaling.
\end{itemize}


\section{Related Work}
\textbf{LLM-based Multi-Agent Systems.}\quad
While agentic capabilities exist on a spectrum, traditional Single-Agent Systems (SAS) operate via a monolithic control loop governed by a single LLM, even when utilizing tools or self-reflection \citep{yao2023react, madaan2023selfrefine}. In contrast, Multi-Agent Systems (MAS) distribute cognitive load across specialized sub-agents with distinct contexts and objectives \citep{xi2023risepotentiallargelanguage}. By interacting through structured communication channels, MAS synthesize individual outputs into cohesive reasoning, inspiring diverse architectures from dynamic routing \citep{su2025toolorchestraelevatingintelligenceefficient,zhang2026routerr} to skill-centric topologies \citep{anthropic_agent_skills_2024}. 

The performance of MAS is fundamentally constrained by orchestration. Earlier methods relied on manual heuristics or inference-time routing~\citep{Ke2025MASZero,zhang2025aflow, liu2024a}, but recent literature has shifted toward \textit{training-time orchestration}. MAS-Orchestra \citep{ke2026masorchestraunderstandingimprovingmultiagent} pioneers training-time orchestration by formulating agent coordination as a holistic, function-calling reinforcement learning problem. As the first framework to generate complete multi-agent topologies in a single step, it achieves state-of-the-art, Pareto-optimal performance across diverse reasoning domains. Furthermore, by abstracting sub-agents as modular functions, it provides a universally representative and highly scalable baseline, making it the ideal foundational framework for our experiments.
However, training this orchestrator relies on sparse, final-answer rewards. Consequently, the system must execute full multi-agent rollouts to receive any feedback, resulting in an immense computational overhead. 
\ours addresses this by utilizing intermediate training artifacts to learn a reward model from automatically constructed preference pairs, enabling direct evaluation of orchestration plans without the prohibitive overhead of exhaustive sub-agent execution.

\textbf{Reward Modeling for LLMs.}\quad
Historically, Reward Modeling (RM) was primarily developed for Single-Agent Systems (SAS) to align LLMs through Reinforcement Learning from Human Feedback (RLHF) ~\citep{ouyang2022training}. In this context, reward models are typically trained as discriminative scoring functions over human-annotated preference pairs, often utilizing a Bradley-Terry optimization objective as in PILAF~\citep{feng2025pilaf}. However, directly applying these single-agent reward models to Multi-Agent Systems introduces severe credit assignment challenges, as interaction histories and prompts vary dynamically across distinct roles and conversation turns~\citep{zhao2026strongermas}. Crucially, as established in MAS-ProVe~\citep{venkataramani2026masproveunderstandingprocessverification},
traditional scalar and process reward models (PRMs) trained on single-agent paths struggle to reliably evaluate the out-of-distribution, conversational dynamics inherent to multi-agent orchestration. 

To address this, recent literature has begun tailoring RM and reinforcement learning explicitly for collaborative MAS. For instance, MARTI ~\citep{zhang2026marti} combines rule-based verifiable rewards with LLM-based generative rewards, and OrchMAS~\citep{feng2026orchmasorchestratedreasoningmulti} trains its orchestration policy using a layered critique paradigm where the rewards are deterministically calculated based on structural completeness and reference-based precision. These frameworks are not applicable as baselines for our setting because they are policy optimization architectures designed to \textit{consume} rewards, rather than methodologies for \textit{training} a reward model. They rely on static external oracles, lacking the structural mechanisms to automatically construct preference pairs from distributed rollouts. In contrast, \ours bridges this gap by natively extracting win-lose pairs from intermediate model checkpoints, optimizing a standalone reward model to provide the scalable evaluative signal required for training-time orchestration.


\section{Methodology}

This section presents our proposed orchestration reward modeling framework, \ours, for improved test-time scaling and training efficiency for MAS orchestrators.
\Secref{sec:method-preliminary} introduces the necessary preliminaries, discussing reward modeling for LLMs and its integration into test-time scaling and training of policy models.
\Secref{sec:method-main} provides an overview of \ours as a complete pipeline including data construction, reward model training, and \Secref{sec:method-application} discusses its application to test-time scaling of MAS and training of orchestrator models.

\subsection{Preliminaries}
\label{sec:method-preliminary}

\textbf{Notation.}\quad
Throughout this paper, scalars are denoted by lowercase letters (e.g., $x$), vectors or token/word sequences by bold lowercase letters (e.g., $\vx$), random vectors by bold lowercase letters with a random-variable notation (e.g., $\rvx$), and matrices or collections of tokens, words, or phrases by bold uppercase letters (e.g., $\mX$). We use $[m]=\{1,2,\ldots,m\}$ to denote the set of consecutive integers from $1$ to $m$.
We follow the convention of using $\basepolicy(\cdot)$, $\policy(\cdot)$, and $\pi_{\mathbf{\Theta}}$ as the base model, being-trained policy model, and the herd of models in MAS, respectively. 

\textbf{Training RMs from Human Preferences.}\quad
Given a prompt $\vx$, candidate responses $(\vy_w, \vy_l)$ are generated and labeled by the base model $\pi_0$ and human annotators, where $(\vy_w, \vy_l)$ indicates the (preferred, dispreferred) pair. 
These preferences are assumed to arise from an underlying latent reward function $r^*(\vx, \vy)$. 
The probability of $\vy_1$ being better than $\vy_2$ is typically modeled by the Bradley-Terry (BT) formulation~\citep{zhang2026bradleyterry}:
\begin{equation}
    p^*(\vy_1 \succ \vy_2 \mid \vx) \triangleq \sigma[r^*(\vx, \vy_1)-r^*(\vx, \vy_2)]. 
\end{equation}
To learn the transferrable capabilities of quality judge of LLM responses, a parametric reward model $r_\vphi(\vx, \vy)$ is trained on a dataset of comparisons $\mathcal{D} = \{\vx^{(i)}, \vy_w^{(i)}, \vy_l^{(i)}\}$ by maximizing likelihood, yielding the standard logistic loss:
\begin{equation}
\gL_{\text{RM}}(\vphi) \triangleq -\E_{\gD} \left[\log \sigma\big(r_\vphi(\vx, \vy_w) - r_\vphi(\vx, \vy_l)\big)\right].
\end{equation}

\begin{figure}[t]
    \centering
    \includegraphics[width=1.0\linewidth]{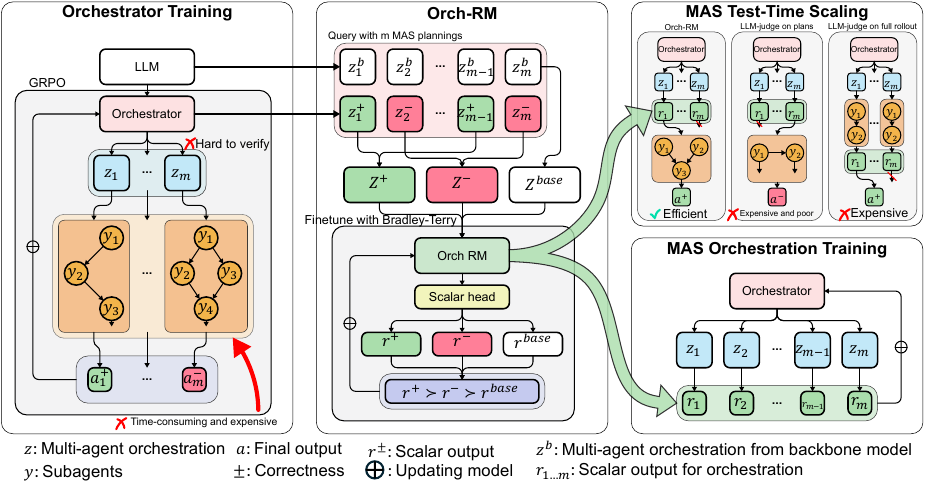}
    \vspace{-0em}
    \caption{
    \textbf{Overview of Orchestration Reward Modeling~(\ours).}
    We construct preference pairs (\emph{``specialized-over-base''} and \emph{``correct-over-incorrect''}) by collecting intermediate artifacts from trained MAS orchestrator models~\cite{ke2026masorchestraunderstandingimprovingmultiagent},   and optimize the reward model with a Bradley-Terry objective. 
    Once trained, \ours can be used for both test-time trajectory selection and continued orchestrator training, where it provides scalar rewards for candidate \emph{orchestrations} without requiring \emph{full sub-agent inference.}
    }
    \label{fig:main}
    \vspace{-0em}
\end{figure}

\textbf{RMs for Improving Policy Models.}\quad
The learned reward model $r_{\vphi}$ then provides scalar feedback for optimizing a policy $\pi_\vtheta$. The typical objective balances reward maximization while staying close to a reference policy $\pi_{0}$ through KL regularization, optimized using reinforcement learning algorithms (such as PPO~\citep{ppo_schulman2017proximal} and GRPO~\citep{shao2024deepseekmath}): 
\begin{equation}
\gL_{\text{RLHF}}(\vtheta) \triangleq - \E_{\vx\sim \gD, \vy \sim \pi_\theta(\cdot\mid \vx)} \big[r_\vphi(\vx, \vy) - \beta D_{\mathrm{KL}}\big[\pi_\vtheta(\vy \mid \vx) \,\|\, \pi_{0}(\vy \mid \vx)\big]\big].
\end{equation}


\textbf{RMs for Test-time Scaling.}\quad
For tasks where the final answers can be effectively aggregated, e.g., math and short-form question answering, scaling up test-time compute by sampling multiple answers in parallel and then aggregating them has shown promise in significantly improving the model performance, e.g., majority voting~\citep{wang2023selfconsistency}. 
An effective reward model can further improve this process by providing a learned scoring function over sampled responses. 
\begin{itemize}[leftmargin=24pt]
\item \emph{Best-of-$N$ (BoN).} BoN selects the single response with the highest reward:
\begin{equation}
    \vy^* = \arg\max\nolimits_{\vy_n} r_\vphi(\vx, \vy_n), \quad \vy_n \sim \pi_\vtheta(\vy \mid \vx),
\end{equation}
which replaces uniform aggregation with reward-based selection, allowing the system to identify higher-quality trajectories even when they are not the most frequent.

\item \emph{Weighted BoN (w-BoN).} Beyond selecting a single trajectory, w-BoN aggregates candidates at the answer level using reward-weighted voting. Let $\mathcal{Y}(\va)$ denote the set of responses that map to answer $\va$. The induced answer distribution becomes
\begin{equation}
    \pi_{\text{w-BoN}}(\va \mid \vx) \propto \sum\nolimits_{\vy_n \in \mathcal{Y}(\va)} \exp\big(r_\vphi(\vx, \vy_n)\big),
\end{equation}
which re-weights each sample according to its reward score. This formulation combines the diversity of multiple samples with the discriminative power of the reward model, often yielding more robust performance than both majority voting and standard BoN.
\end{itemize}

The following section will discuss the complications and challenges of integrating the idea of reward modeling for multi-agent orchestration.

\subsection{\ours: Reward Modeling for Multi-Agent Orchestration}
\label{sec:method-main}

\textbf{Orchestration-Based LLM Multi-Agent Systems.}\quad
Given an input query $\vx$, an orchestration-based MAS first uses a policy model $\policy$ to generate a multi-agent orchestration $\vz$. 
The orchestration specifies the sub-agent roles, role descriptions, collaboration structure, and interaction graph among agents. 
It is then executed by the sub-agents, producing a sequence of intermediate responses $\mY=\{\vy_1, \vy_2, \cdots, \vy_k\}$ following the workflow order. 
After all sub-agents complete inference, the system aggregates their outputs to produce a final answer $\va$. 
We denote a complete execution as a trajectory
$
\traj = (\vx, \vz, \mY, \va).
$
Each trajectory is assigned an outcome label based on the correctness of $\va$, which is later used to construct reward model training data.

\textbf{Motivation and Overview.}\quad
From the formulation of a MAS trajectory $\traj$, we identify two key bottlenecks for effective MAS:
\begin{itemize}[leftmargin=24pt]
    \item \emph{Performance bottleneck}, determined by the quality of the orchestration $\vz$. Prior work shows that well-designed orchestration can substantially improve MAS effectiveness~\citep{ke2026masorchestraunderstandingimprovingmultiagent}. 
    \item \emph{Efficiency bottleneck}, driven by the cost of executing sub-agent rollouts $\mY$, compounded by the fact that even an intermediate response $\vy_1$ requires the execution of multiple agents. These executions are time-consuming, limiting the scalability of both inference and training. In particular, supervision for orchestration quality is costly, as it typically requires executing $\vz$ to completion.
\end{itemize}
To address these challenges, we aim to learn a reward model $\rmModel(\vx, \vz)$ that assigns a scalar score to an orchestration $\vz$, directly estimating its quality. Crucially, this model operates at the orchestration level rather than over full trajectories. This design offers two advantages: (i) at inference time, it enables early-stage intervention in MAS execution, leading to more token-efficient test-time scaling; (ii) during training, it provides dense feedback for the orchestrator without requiring expensive full trajectory rollouts. 
\Figref{fig:main} illustrates the training pipeline of \ours and its downstream applications, which we describe in the following sections.


\textbf{Data Construction.}\quad 
To the best of our knowledge, there is no existing labeled dataset exclusively for MAS orchestration.
As discussed earlier, collecting such data faces two main challenges: (1) Human annotations may be misaligned with machine objectives, as humans often emphasize readability or interpretability rather than execution effectiveness, introducing noise. (2) Labeling orchestration quality is expensive, as it typically requires executing full MAS trajectories.

To address these challenges, we leverage an off-the-shelf trained MAS orchestrator and the intermediate artifacts generated during its training~\citep{ke2026masorchestraunderstandingimprovingmultiagent}. Specifically, we construct two sources of orchestration comparisons for training \ours: 
\begin{itemize}[leftmargin=24pt]
    \item \emph{Specialized over base} ($\policy \succ \basepolicy$): we compare trajectories generated by the trained orchestrator $\policy$ against those from the backbone model $\basepolicy$, where the former generally produces higher-quality orchestrations due to task-specific optimization \footnote{Repeated-execution win rates for SoB range from 94.0\% to 96.5\% across three datasets (see \appref{app:preference_reliability})}.
    
    \item \emph{Correct over incorrect} ($\vz^+ \succ \vz^-$): for each input, we use sampled trajectories from $\policy$ and label them based on the correctness of their final outputs, treating correct trajectories (\chosentraj) as preferred over incorrect ones (\rejectedtraj). Notably, this source is constructed entirely from accessible rollout artifacts \emph{without additional annotation} 
    \footnote{
    Repeated-execution win rates for CoI range from 67.5\% to 72.5\% across three datasets (see \appref{app:preference_reliability}).}.
\end{itemize}

We combine these two sources of pairwise data using a fixed SoB:CoI mixing ratio of $1{:}3$ to train the reward model. 
For an ablation study on the effect of data mixing ratios in \ours training, please refer to \Secref{sec:experiment-tts}.

\textbf{Training Objective.}\quad 
We train $\rmModel$ to assign higher scores to higher-quality orchestration than to lower-quality orchestration, using the standard Bradley-Terry modeling as detailed in \Secref{sec:method-preliminary}: 
\begin{equation}
    \gL_{\text{\ours}}(\vphi) \triangleq -\E_{\gD} \left[\log \sigma\big(r_\vphi(\vx, \vz_w) - r_\vphi(\vx, \vz_l)\big)\right].
\end{equation}




\subsection{Applying \ours to MAS: Efficient Test-Time Scaling and Orchestrator Training}
\label{sec:method-application}

\textbf{MAS Test-Time Scaling}.\quad
We perform test-time scaling by sampling multiple orchestrations from the orchestrator policy $\policy$ and using the reward model $\rmModel$ to score them. Specifically, given an input $\vx$, we sample a set of orchestrations $\mZ \triangleq \{\vz_{n}\}_{n\in[N]}$ of size $N$ in parallel and select the highest-scoring orchestration $\hat{\vz}$ with our trained orchestration reward model $\rmModel$: 
\begin{equation}
\hat{\vz} = \arg\max\nolimits_{\vz\in \mZ} \left\{\rmModel(\vx, \vz)\right\}.
\end{equation}
The selected best orchestration is then used for the full MAS trajectory generation $\mY\sim \pi_{\mathbf{\Theta}}(\cdot\mid \vx, \hat{\vz})$.

Unlike typical MAS test-time scaling~\cite{venkataramani2026masproveunderstandingprocessverification} that repeatedly executes diverse sub-agent behaviors in parallel, our \ours operates at the \emph{orchestration level}, directly \emph{evaluating high-level plans before any sub-agent execution}. This avoids expensive, redundant sub-agent rollouts at inference time and enables much more efficient scaling, as the reward model captures the quality of both the orchestration strategy and the induced sub-agent behaviors. 

\textbf{Orchestrator Training}.\quad
Beyond inference, we integrate $\rmModel$ into the training of the orchestration policy. We consider both continued training from a pretrained $\policy$ and training from scratch initialized from the backbone $\basepolicy$. In both settings, the reward model replaces the original task-level reward by assigning a scalar score to each orchestration. 
For a group of sampled orchestrations $\{\vz_{n}\}_{n\in[N]}$, we compute normalized advantages with
\begin{equation}
    A_{n} = \rmModel(\vx, \vz_n) - \frac{1}{N} \sum\nolimits_{j=1}^{N} \rmModel(\vx, \vz_j),
\end{equation}
and optimize the policy using the GRPO objective. Importantly, this formulation provides supervision at the orchestration level, allowing the model to implicitly learn better coordination of sub-agent behaviors; it \emph{avoids expensive sub-agent execution (rollouts)} and does not require explicit annotations for each step. As a result, training becomes more efficient and scalable, while still improving overall system performance.

\section{Experiments}
We evaluate \ours through extensive experiments. 
\Secref{sec:experiment-setup} presents the experimental setup. 
In \Secref{sec:experiment-tts}, we first evaluate the application of \ours to orchestration-level MAS test-time scaling, followed by an ablation study on the data mixing ratios used for reward model training. 
Finally, \Secref{sec:experiment-training} studies the effectiveness of using \ours as the reward signal for MAS orchestrator training, under both training-from-scratch and continued-training settings.

\subsection{Experimental Setup}
\label{sec:experiment-setup}

\textbf{Models.}\quad
We follow the setup of MAS-Orchestra~\citep{ke2026masorchestraunderstandingimprovingmultiagent}, using \texttt{Qwen2.5-7B-Instruct}~\citep{qwen2.5} as the orchestration policy $\policy$ and \texttt{GPT-OSS-120B}~\citep{agarwal2025gpt} as the backbone model for all sub-agent calls during MAS trajectory generation. For orchestration-level reward modeling, we initialize from \texttt{Skywork-Reward-Llama-3.1-8B}~\citep{liu2024skywork} and fine-tune it on the constructed preference data.

\textbf{Benchmarks.}\quad
We base our evaluation on three main benchmarks widely adopted in prior works~\citep{ke2026masorchestraunderstandingimprovingmultiagent}: \textbf{AIME 24\&25}~\citep{aime} for challenging mathematical reasoning with moderate MAS requirements, \textbf{BrowseComp+}~\citep{chen2025BrowseCompPlus} for web-based QA requiring high-level planning and precise tool use, and \textbf{HotpotQA}~\citep{yang-etal-2018-hotpotqa} for multi-hop reasoning. To assess cross-domain robustness, we additionally report out-of-distribution results on \textbf{GPQA}~\citep{rein2023gpqagraduatelevelgoogleproofqa} for scientific reasoning. Further dataset details are provided in \appref{app:dataset_details}


\textbf{Baselines.}\quad
We compare \ours against strong baselines under the same evaluation protocol (see \appref{app:baseline_details} for implementation details).
\begin{itemize}[leftmargin=24pt]
    \item \emph{For test-time scaling:} we include state-of-the-art automatic verification methods for LLMs, applied at both the orchestration level and the full MAS trajectory level when applicable. These include 
    (i) \textbf{general-purpose LLM-as-a-Judge}~\citep{llm-as-a-judge} methods, with base models \texttt{GPT-4.1-mini}, \texttt{GPT-5-mini}, \texttt{GPT-5.2}, and \texttt{GPT-5.4-mini}~\citep{singh2026openaigpt5card}; 
    (ii) \textbf{off-the-shelf reward signals}, including log-probability as internal confidence ($\log P$)~\citep{kumar-etal-2024-confidence} and Skywork-Reward models (\texttt{Llama-3.1-8B}~\citep{liu2024skywork}; \texttt{V2-Llama-3.2-3B}, \texttt{V2-Qwen3-4B}, and \texttt{V2-Qwen3-8B}~\citep{liu2025skywork}); 
    (iii) \textbf{test-time scaling baselines}, including no reward model, oracle selection (Pass@N), and Majority Vote~\citep{wang2023selfconsistency}. 

    \item \emph{For orchestrator training:} we include standard post-training methods for LLMs, including Rejection Sampling Fine-Tuning (\textbf{RFT})~\citep{yuan2023rft} and continued reinforcement learning with Group Relative Policy Optimization (\textbf{GRPO})~\citep{shao2024deepseekmath} applied at the full MAS trajectory level using final correctness as reward~\cite{ke2026masorchestraunderstandingimprovingmultiagent}. We also include orchestration-level training with \textbf{LLM-as-a-Judge}~\citep{llm-as-a-judge} as the reward signal~(with \texttt{GPT-5-mini} as the judge model), and Direct Preference Optimization (\textbf{DPO})~\citep{Rafailov2023dpo} trained on the same static dataset as our \ours.
\end{itemize}

\begin{table}[t]
\centering
\caption{
\textbf{Performance of Orchestration-Level Automatic Verification Methods for MAS Test-Time Scaling.}
We compare \ours primarily against orchestration-level verification methods (bottom two blocks). 
Trajectory-level baselines are included for reference, including MAS-Orchestra~\cite{ke2026masorchestraunderstandingimprovingmultiagent} without verification as a lower bound (LB), and an oracle reward model that always selects the best answer as an upper bound (UB).
\textbf{Acc.:} Best-of-$N$ accuracy ($N{=}8$ across all benchmarks), reported in percentages.
\textbf{\#Tok (M):} Total number of tokens (input + output) used for verification.
\textbf{Boldface} and \underline{underline} indicate the best and the second best results among orchestration-level methods.
}
\label{tab:test-time-selection}
\vspace{-0em}

\resizebox{\textwidth}{!}{%
\setlength{\tabcolsep}{4pt}
\begin{tabular}{lc cc cc cc cc}
\toprule[0.12em]
\multirow{2}{*}{\textbf{Method}}
& \multirow{2}{*}{\textbf{\makecell{Verification\\Level}}}
& \multicolumn{2}{c}{\textbf{AIME 24\&25}}
& \multicolumn{2}{c}{\textbf{BrowseComp+}}
& \multicolumn{2}{c}{\textbf{HotpotQA}}
& \multicolumn{2}{c}{\textbf{GPQA}} \\
\cmidrule(lr){3-4}
\cmidrule(lr){5-6}
\cmidrule(lr){7-8}
\cmidrule(lr){9-10}
& 
& \textbf{Acc.} 
& \textbf{\#Tok (M)}
& \textbf{Acc.} 
& \textbf{\#Tok (M)}
& \textbf{Acc.} 
& \textbf{\#Tok (M)}
& \textbf{Acc.} 
& \textbf{\#Tok (M)}
\\
\midrule

\multicolumn{10}{l}{\textit{Baseline}}
\\
\midrule

\quad MAS-Orchestra~\cite{ke2026masorchestraunderstandingimprovingmultiagent} (LB)
& -
& 63.33 & -
& 9.50 & -
& 42.50 & -
& 63.63 & -
\\

\quad Maj. Vote~\citep{wang2023selfconsistency}
& Trajectory
& 63.33 & -
& 9.50 & -
& 42.50 & -
& 63.63 & -
\\

\quad \texttt{GPT-5-mini}~\citep{singh2026openaigpt5card}
& Trajectory
& 70.00 & 12.62
& 12.50 & 142.80
& 44.00 & 14.86
& 68.89 & 29.76
\\

\quad Oracle (UB)
& Trajectory
& 76.67 & -
& 24.50 & -
& 57.00 & -
& 84.85 & -
\\

\midrule

\multicolumn{10}{l}{\textit{LLM-as-a-Judge}}
\\
\midrule

\quad \texttt{GPT-4.1-mini}~\citep{openai2025_gpt41_nano}
& Orchestration
& 58.33 & 2.52
& 9.00 & \underline{8.63}
& 42.00 & \underline{7.66}
& 67.68 & 7.85
\\

\quad \texttt{GPT-5-mini}~\citep{singh2026openaigpt5card}
& Orchestration
& 65.00 & 2.53
& 10.50 & 8.66
& 40.50 & 7.69
& \textbf{71.72} & 7.88
\\

\quad \texttt{GPT-5.2}~\citep{singh2026openaigpt5card}
& Orchestration
& 65.00 & \underline{2.40}
& 10.00 & 8.75
& 42.00 & 7.77
& 68.02 & 7.89
\\

\quad \texttt{GPT-5.4-mini}~\citep{singh2026openaigpt5card}
& Orchestration
& \underline{66.67} & 2.53
& 11.00 & 8.72
& \textbf{46.00} & 7.75
& 68.13 & \underline{7.83}
\\

\midrule

\multicolumn{10}{l}{\textit{Reward Model}}
\\
\midrule

\quad $\log P$~\citep{kumar-etal-2024-confidence}
& Orchestration
& 55.00 & -
& 10.50 & -
& 39.50 & -
& \underline{71.21} & -
\\

\quad \texttt{Skywork-Llama-3.1-8B}~\cite{liu2024skywork}
& Orchestration
& 61.67 & \textbf{2.38}
& 11.50 & \textbf{8.26}
& 40.50 & \textbf{7.32}
& 64.67 & 7.49
\\

\quad \texttt{Skywork-V2-Llama-3.2-3B}~\cite{liu2025skywork}
& Orchestration
& 58.33 & \textbf{2.38}
& \underline{12.00} & \textbf{8.26}
& 41.50 & \textbf{7.32}
& 64.65 & 7.49
\\

\quad \texttt{Skywork-V2-Qwen3-4B}~\cite{liu2025skywork}
& Orchestration
& 61.67 & \textbf{2.38}
& 8.00 & \textbf{8.26}
& 42.00 & \textbf{7.32}
& 68.69 & 7.49
\\

\quad \texttt{Skywork-V2-Qwen3-8B}~\cite{liu2025skywork}
& Orchestration
& 60.00 & \textbf{2.38}
& 9.00 & \textbf{8.26}
& 41.00 & \textbf{7.32}
& 68.18 & 7.49
\\

\rowcolor{lightergray}
\quad \ours~(Ours)
& Orchestration
& \textbf{68.33} & \textbf{2.38}
& \textbf{14.00} & \textbf{8.26}
& \underline{42.50} & \textbf{7.32}
& 66.16 & \textbf{7.49}
\\

\bottomrule[0.12em]
\end{tabular}
}
\vspace{-0em}
\end{table}

\textbf{Implementation Details.}\quad
We strictly follow the setup of MAS-Orchestra~\cite{ke2026masorchestraunderstandingimprovingmultiagent} and train domain-specific reward models using corresponding data and off-the-shelf orchestrator models. 
For detailed dataset description, please refer to \appref{app:dataset_details}. 
\begin{itemize}[leftmargin=24pt]
    \item For \emph{``specialized-over-base''} pairs, we sample 10 \emph{orchestrations} per query from both the orchestration policy $\policy$ and the base policy $\basepolicy$; 
    \item For \emph{``correct-over-incorrect''} pairs, we collect \emph{full trajectories} from MAS-Orchestra training logs (up to the final step)\footnote{
        We follow the setting of MAS-Orchestra~\cite{ke2026masorchestraunderstandingimprovingmultiagent} as it provides the model checkpoints and full training log which are available at \url{https://github.com/SalesforceAIResearch/MAS-Orchestra}.
    }
    and pair orchestrations that lead to correct versus incorrect outcomes.
\end{itemize}
We combine these two types of pairwise data at a Specialized-over-Base (SoB) to Correct-over-Incorrect (CoI) ratio of 1:3 as it provides the best generalization ability for the model,
yielding 4 \emph{``specialized-over-base''} pairs and 12 \emph{``correct-over-incorrect''} pairs per query. Ablation results on the effect of this mixing ratio are provided in \Secref{sec:experiment-tts}.

\subsection{\ours for MAS Test-Time Scaling}
\label{sec:experiment-tts}

\Tabref{tab:test-time-selection} reports the performance of \ours for MAS test-time scaling against other verification baselines across four datasets. \Tabref{tab:ablation_rm_data} further analyzes the effect of different data mixing ratios used in \ours training. \Tabref{tab:score_reliability} shows the correlation between \ours scores and empirical orchestration success rates estimated from repeated executions of fixed plans.


\textbf{\ours improves the accuracy-efficiency trade-off for MAS test-time scaling.}\quad 
As shown in Table~\ref{tab:test-time-selection}, \ours achieves the strongest overall accuracy-efficiency trade-off among orchestration-level verification methods. Compared to the self-consistency baseline, \ours improves Best-of-$N$ accuracy from 63.33\% to 68.33\% on AIME 24\&25 and from 9.50\% to 14.00\% on BrowseComp+, while using only 2.38M and 8.26M verification tokens, respectively. \ours also outperforms all orchestration-level LLM-as-a-Judge and off-the-shelf reward model baselines on these two benchmarks. In particular, on BrowseComp+, \ours surpasses the strongest trajectory-level LLM judge (12.50\% by GPT-5-mini) while reducing verification cost from 142.80M to 8.26M tokens.
Additionally, as shown in \Figref{fig:fig1}, \ours continues to benefit as the parallel sampling size scales up to 64 on AIME 24\&25, demonstrating the scalability of its reward signals for large-scale test-time sampling.

\textbf{\ours~benefits from combining multiple sources of orchestration preference data.}\quad
Table~\ref{tab:ablation_rm_data} shows that the data mixing ratio substantially affects reward-guided test-time scaling performance. Using only Correct-over-Incorrect (\emph{CoI}) pairs or only domain-level comparisons leads to weaker results than the default mixed setting. In contrast, the default configuration in row (e), which combines Specialized-over-Base (\emph{SoB}) and Correct-over-Incorrect (\emph{CoI}) pairs with a 1{:}3 ratio, achieves the best overall performance, reaching 68.33\% accuracy on AIME 24\&25 and 14.00\% on BrowseComp+. This setting also produces strong average normalized advantages (0.25 on AIME and 0.31 on BrowseComp+), indicating improved separation between correct and incorrect orchestrations.

\begin{table}[t]
\centering
\caption{
\textbf{Ablation Study on Data Mixing Ratios for \ours Training.} 
The evaluation setup follows \Tabref{tab:test-time-selection}. 
We consider three sources of pairwise data for orchestration reward model training: (i) Specialized-over-Base (\emph{SoB}), (ii) Specialized-over-Other-domains (\emph{SoO}), and (iii) Correct-over-Incorrect (\emph{CoI}). 
\textbf{Avg. Adv.:} average normalized advantage between orchestrations leading to correct and incorrect answers.
\textbf{(a):} untrained \texttt{Skywork-Reward-Llama-3.1-8B} baseline;
\textbf{(b)-(d):} alternative data mixing ratios for \ours; 
\textbf{(e):} default setting of \ours;
\textbf{(f):} oracle reward model upper bound.
\textbf{Boldface} denotes the best performance.
}
\label{tab:ablation_rm_data}
\vspace{-0em}

\resizebox{1\textwidth}{!}{%
\setlength{\tabcolsep}{5pt}
\begin{tabular}{lcc ccc cc cc cc }
\toprule[0.12em]
& \multirow{2}{*}{\textbf{Method}}
& \multirow{2}{*}{\textbf{\makecell{Verification\\Level}}}
& \multicolumn{3}{c}{\textbf{Data Mix Ratio}}
& \multicolumn{2}{c}{\textbf{AIME 24\&25}}
& \multicolumn{2}{c}{\textbf{BrowseComp+}}
& \multicolumn{2}{c}{\textbf{HotpotQA}} \\

\cmidrule(lr){4-6}
\cmidrule(lr){7-8}
\cmidrule(lr){9-10}
\cmidrule(lr){11-12}
& & 
& \emph{SoB}
& \emph{SoO}
& \emph{CoI}
& \emph{Acc.}
& \emph{Avg. Adv.}
& \emph{Acc.}
& \emph{Avg. Adv.}
& \emph{Acc.}
& \emph{Avg. Adv.}
\\
\midrule




(a)
& \multirow{5}{*}{\ours}
& \multirow{5}{*}{Orchestration}
& - & - & - 
& 61.67 & \textbf{0.32}
& 11.50 & 0.12
& 40.50 & \textbf{0.03}
\\

(b)
& 
& 
& 3 & 2 & 0
& 63.33 & 0.18
& \textbf{14.00} & 0.18
& 41.00 & -0.05
\\

(c)
& & 
& 0 & 0 & 1
& 63.33 & 0.20
& 10.50 & 0.22
& 42.00 & -0.04
\\

(d)
& &
& 1 & 0 & 1
& 63.33 & 0.30
& 11.00 & \textbf{0.32}
& \textbf{42.50} & -0.03
\\

(e)
& &
& \cellcolor{lightergray}1 
& \cellcolor{lightergray}0 
& \cellcolor{lightergray}3
& \cellcolor{lightergray}\textbf{68.33} & \cellcolor{lightergray}0.25
& \cellcolor{lightergray}\textbf{14.00} & \cellcolor{lightergray}0.31
& \cellcolor{lightergray}\textbf{42.50} & \cellcolor{lightergray}-0.03
\\

\midrule

(f)
& Oracle (UB)
& Trajectory
& - & - & - 
& 76.67 & -
& 24.50 & -
& 57.00 & -
\\

\bottomrule[0.12em]
\end{tabular}
}
\vspace{-1em}
\end{table}

    

\begin{wraptable}{r}{0.44\textwidth}
    \vspace{-\intextsep}
    \setlength{\abovecaptionskip}{0pt}
    \centering
    \small

    \caption{Spearman correlation between Orch-RM scores and empirical plan success rates, estimated from five executions of each fixed orchestration plan.}
    \label{tab:score_reliability}

    \setlength{\tabcolsep}{3pt}
    \begin{tabularx}{\linewidth}{
        @{}>{\raggedright\arraybackslash}X
        >{\centering\arraybackslash}X@{}
    }
        \toprule
        \textbf{Benchmark}
        & \textbf{\shortstack{Spearman correlation}} \\
        \midrule
        AIME 24\&25       & 0.724 \\
        BrowseComp+ & 0.597 \\
        HotpotQA    & 0.546 \\
        \bottomrule
    \end{tabularx}
    \vskip -0.3cm
\end{wraptable}

\textbf{\ours scores correlate with orchestration reliability.}\quad
We operationalize orchestration quality as the probability that executing a plan produces a correct final answer. To estimate this quantity beyond a single execution outcome, we fix each orchestration plan and repeat its sub-agent execution five times. We use the proportion of correct executions as an empirical estimate of the plan's expected accuracy, and compute Spearman's rank correlation between these estimates and the scores assigned by \ours. As shown in Table~\ref{tab:score_reliability}, the correlations are positive across all three benchmarks: 0.724 on AIME 24\&25, 0.597 on BrowseComp+, and 0.546 on HotpotQA. These results suggest that \ours assigns higher scores to plans that succeed more reliably across stochastic executions, rather than merely reproducing the outcome of a single rollout. This analysis supports alignment with execution reliability, although it does not disentangle orchestration quality from answer correctness, which remains our operational measure.


In both \Tabref{tab:test-time-selection} and \Tabref{tab:ablation_rm_data}, the gains are smaller on HotpotQA and GPQA, where 
our diversity analysis (\appref{app:orchestration_diversity}) reveals highly homogeneous orchestration structures on these benchmarks, which may partly explain the limited benefits of orchestration-level verification (see \appref{sec:appendix_rollouts} for qualitative examples).
In particular, on HotpotQA, all settings achieve similar accuracy (40.50-42.50\%) and slightly negative normalized advantages, suggesting that many trajectories share similar orchestration structures and are therefore difficult to distinguish at the orchestration level. Nevertheless, \ours remains competitive while maintaining substantially lower verification cost than trajectory-level methods.\footnote{For plans generated by GPT-5-mini rather than MAS-Orchestra, \ours matches or exceeds the LLM-as-a-Judge baseline across three benchmarks when scoring their XML representations. See \appref{app:format_transfer} for details.} Overall, these results suggest that combining correctness-based and domain-level orchestration comparisons provides the most effective supervision signal for \ours.

\subsection{\ours for MAS Orchestrator Training}
\label{sec:experiment-training}



\Tabref{tab:training} reports the performance of using \ours as the orchestration-level reward signal for MAS orchestrator training under two settings: \emph{(i) training from scratch} and \emph{(ii) continued training}. \Figref{fig:post-training} further illustrates the performance-efficiency trade-off of different post-training methods during continued training of the MAS orchestrator. We additionally compare the estimated computational costs of continued training in \Tabref{tab:training_flops_summary}, accounting for upfront reward-model training and subsequent orchestrator optimization.

\textbf{\ours enables efficient MAS orchestrator training from scratch.}\quad
As shown in Table~\ref{tab:training}, training an orchestrator from scratch with \ours substantially improves over the untrained base model across all benchmarks, increasing majority-vote accuracy from 23.83\% to 62.67\% on AIME 24\&25 and from 0.35\% to 4.40\% on BrowseComp+. Despite relying only on orchestration-level supervision, \ours achieves performance competitive with trajectory-level MAS-Orchestra training.

\begin{wraptable}{r}{0.44\textwidth}
    \vspace{-\intextsep}
    \setlength{\abovecaptionskip}{0pt}
    \centering
    \small

    \caption{Estimated computational cost of continued orchestrator training at $N=8$.Ratios are total GRPO FLOPs over total \ours FLOPs, including upfront reward-model training.}
    \label{tab:training_flops_summary}

    \setlength{\tabcolsep}{3pt}
    \begin{tabularx}{\linewidth}{
        @{}>{\raggedright\arraybackslash}X
        >{\centering\arraybackslash}X@{}
    }
        \toprule
        \textbf{Training dataset}
        & \shortstack{\textbf{FLOPs ratio}\\(GRPO / \ours)} \\
        \midrule
        DeepScaleR  & $9.3\times$ \\
        BrowseComp+ & $64.7\times$ \\
        HotpotQA    & $3.2\times$ \\
        \bottomrule
    \end{tabularx}
    \vspace{-\intextsep}
\end{wraptable}

\textbf{\ours significantly improves the performance-efficiency trade-off for continued orchestrator training.}\quad
Under continued training, \ours achieves the best overall performance on AIME 24\&25 and BrowseComp+, reaching $69.83\%$ and $10.90\%$ majority-vote accuracy, respectively. 
Crucially, \textit{\ours is the only method that consistently improves performance under the continued training setting,} demonstrating the reliability of its reward signals.
As shown in \Figref{fig:post-training}, \ours consistently provides a stronger performance-efficiency trade-off than trajectory-level post-training methods such as RFT and GRPO. In particular, on BrowseComp+, \ours improves majority-vote accuracy from $8.90\%$ to $10.90\%$ while using over $10\times$ fewer training tokens than GRPO. These results highlight the efficiency advantage of orchestration-level reward supervision, which avoids expensive full-trajectory optimization over sub-agent rollouts.

\begin{table}[t]
\centering
\caption{
\textbf{Performance of \ours for MAS Orchestrator Training.}
Values are mean $\pm$ standard deviation over 10 bootstrap resamples of evaluation questions.
\textbf{(a):} base orchestrator without additional training;
\textbf{(b)--(c):} training-from-scratch methods initialized from the base model;
\textbf{(d)--(h):} continued-training methods initialized from the off-the-shelf MAS-Orchestra model.
\textbf{Boldface} and \underline{underline} denote the best and second-best mean performance in each column, respectively. Tied means receive the same marking.
}
\label{tab:training-comparison-all-datasets}
\label{tab:training}
\vspace{-0.5em}
\setlength{\tabcolsep}{4pt}

\resizebox{\textwidth}{!}{
\begin{tabular}{llccccccccc}
\toprule
& \multirow{2}{*}{\textbf{Method}}
& \multirow{2}{*}{\textbf{\makecell{Sampling\\Level}}}
& \multicolumn{2}{c}{\textbf{AIME 24\&25}}
& \multicolumn{2}{c}{\textbf{BrowseComp+}}
& \multicolumn{2}{c}{\textbf{HotpotQA}}
& \multicolumn{2}{c}{\textbf{GPQA}} \\
\cmidrule(lr){4-5}
\cmidrule(lr){6-7}
\cmidrule(lr){8-9}
\cmidrule(lr){10-11}
& &
& \textbf{Acc.} & \textbf{Maj. Vote}
& \textbf{Acc.} & \textbf{Maj. Vote}
& \textbf{Acc.} & \textbf{Maj. Vote}
& \textbf{Acc.} & \textbf{Maj. Vote} \\
\midrule

(a)
& \texttt{Qwen2.5-7B-Instruct}~\citep{qwen2.5}
& -
& $30.08 \pm 1.09$ & $23.83 \pm 0.81$
& $2.96 \pm 0.45$ & $0.35 \pm 0.47$
& $31.48 \pm 1.01$ & $34.65 \pm 1.25$
& $24.92 \pm 0.32$ & $19.39 \pm 0.54$ \\
\midrule

\rowcolor{lightergray}
(b)
& \quad + \ours~(Ours)
& Orchestration
& $64.15 \pm 0.80$ & $62.67 \pm 0.86$
& $6.04 \pm 0.48$ & $4.40 \pm 0.52$
& $40.44 \pm 1.04$ & $41.85 \pm 1.38$
& $65.20 \pm 0.47$ & $63.38 \pm 1.02$ \\

(c)
& \quad + MAS-Orchestra~\citep{ke2026masorchestraunderstandingimprovingmultiagent}
& Trajectory
& $59.92 \pm 0.63$ & $62.50 \pm 1.18$
& $\underline{8.93 \pm 0.51}$
& $\underline{8.90 \pm 0.70}$
& $40.43 \pm 0.95$ & $41.80 \pm 1.23$
& $\mathbf{66.16 \pm 0.70}$ & $64.34 \pm 0.26$ \\
\midrule

(d)
& \qquad + DPO~\citep{Rafailov2023dpo}
& -
& $63.98 \pm 1.28$
& $\underline{64.83 \pm 1.46}$
& $8.64 \pm 0.31$ & $7.35 \pm 0.34$
& $40.10 \pm 1.14$ & $40.10 \pm 1.10$
& $64.54 \pm 0.39$ & $64.49 \pm 0.24$ \\

(e)
& \qquad + RFT~\citep{yuan2023rft}
& Trajectory
& $\underline{64.33 \pm 0.84}$ & $63.83 \pm 0.81$
& $6.91 \pm 0.59$ & $5.50 \pm 0.53$
& $39.07 \pm 0.73$ & $41.05 \pm 0.76$
& $64.75 \pm 0.36$ & $64.14 \pm 0.71$ \\

(f)
& \qquad + GRPO~\citep{shao2024deepseekmath}
& Trajectory
& $63.78 \pm 1.06$
& $\underline{64.83 \pm 0.53}$
& $6.48 \pm 0.38$ & $5.70 \pm 0.54$
& $\underline{40.95 \pm 1.01}$
& $\mathbf{43.28 \pm 1.22}$
& $65.58 \pm 0.66$
& $\underline{65.25 \pm 0.40}$ \\

(g)
& \qquad + LLM-as-a-Judge~\citep{llm-as-a-judge}
& Orchestration
& $59.13 \pm 1.39$ & $61.17 \pm 0.81$
& $8.53 \pm 0.39$ & $8.13 \pm 0.28$
& $39.69 \pm 0.81$ & $42.00 \pm 0.91$
& $61.48 \pm 0.44$ & $61.26 \pm 0.54$ \\

\rowcolor{lightergray}
(h)
& \qquad + \ours~(Ours)
& Orchestration
& $\mathbf{67.72 \pm 1.44}$
& $\mathbf{69.83 \pm 0.53}$
& $\mathbf{9.75 \pm 0.51}$
& $\mathbf{10.90 \pm 0.70}$
& $\mathbf{42.08 \pm 0.85}$
& $\underline{42.10 \pm 0.99}$
& $\underline{66.07 \pm 0.90}$
& $\mathbf{66.11 \pm 0.73}$ \\

\bottomrule
\end{tabular}
}
\end{table}

\begin{figure}[t]
    \centering
    \includegraphics[width=1\linewidth]{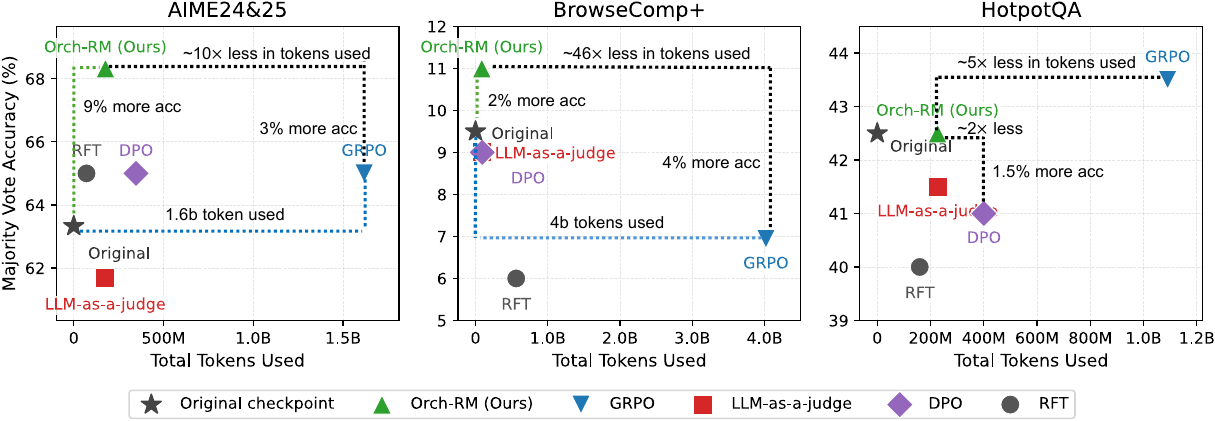}
    \vspace{-1em}
    \caption{
    \textbf{Continued Training of MAS Orchestrators with Different Reward Signals.}
    We compare the performance-efficiency trade-off of different post-training methods by reporting downstream majority-vote accuracy and the total number of training tokens consumed. Across AIME 24\&25 and BrowseComp+, \ours achieves the strongest overall trade-off, consistently improving over the original MAS-Orchestra~\cite{ke2026masorchestraunderstandingimprovingmultiagent} checkpoint while using substantially fewer training tokens than trajectory-level reinforcement learning methods such as GRPO. In particular, \ours attains comparable or stronger performance using up to $\sim$10$\times$ fewer tokens on AIME 24\&25 and $\sim$46$\times$ fewer tokens on BrowseComp+. 
    }
    \label{fig:post-training}
    \vspace{-1em}
\end{figure}

\textbf{\ours reduces computational cost even after accounting for reward-model training.}\quad
To complement the token-based comparison in \Figref{fig:post-training}, we estimate the computational cost of continued orchestrator training, including upfront reward-model training, orchestrator sampling and optimization, and reward computation. Under the matched $N=8$ setting, Table~\ref{tab:training_flops_summary} shows that GRPO requires approximately 9.3, 64.7, and 3.2 times the FLOPs of \ours on DeepScaleR, BrowseComp+, and HotpotQA, respectively. The largest reduction occurs on BrowseComp+, where full sub-agent execution is particularly costly. These estimates indicate that the computational benefits persist after accounting for reward-model training. Detailed cost breakdowns and estimation assumptions are provided in \appref{app:training_cost}.


\section{Conclusion}
\label{sec:conclusion}

In this work, we propose \ours, an orchestration reward modeling framework for multi-agent LLM systems. By learning orchestration-level reward signals from accessible training artifacts and pairwise orchestration comparisons, \ours avoids costly human annotations and expensive full-trajectory supervision over sub-agent rollouts {during orchestrator inference and training}. We show that \ours can be effectively applied to both MAS test-time scaling and orchestrator training, consistently improving the performance-efficiency trade-off while operating entirely at the orchestration level. Experiments across mathematical reasoning, web-based QA, multi-hop reasoning, and scientific reasoning suggest that orchestration-level reward modeling is a promising future direction for improving automated MAS orchestration.

\textbf{Limitations.}\quad
\ours has several limitations. First, although it avoids expensive human annotations by leveraging existing MAS orchestration artifacts, its performance is still constrained by the diversity and quality of available orchestrator models and sampled trajectories. Constructing larger-scale and more diverse orchestration datasets remains an important challenge. Second, \ours is currently trained in a domain-specific manner, limiting its generalizability. Whether orchestration reward models can benefit from multi-domain and multi-task training remains underexplored.

\begin{ack}
We thank all reviewers, the area chair, and the senior area chair for their
valuable comments.

\textbf{Funding.}
HW is supported by the Amazon Faculty Research Award, Microsoft AI \& Society
Fellowship, Nokia Faculty Research Award, NSF CAREER Award IIS-2340125,
NIH grant R01CA297832, and NSF grant IIS-2127918.

\textbf{Competing interests.}
The authors declare no competing interests.
\end{ack}

{
\bibliography{ref,mas_r1,mas_zero}
\bibliographystyle{abbrv}
}

\appendix
\clearpage
\section*{\LARGE Appendix}
\markboth{Appendix}{Appendix}

In \appref{app:limitations}, we discuss the limitations of our current framework in detail. 
Next, in \appref{app:exp_settings}, we provide implementation details, including baseline descriptions, dataset specifications, and prompt templates. In \appref{app:additional_analyses}, we present additional analyses, covering the reliability of the constructed preferences, orchestration diversity across benchmarks, the computational cost of continued orchestrator training, and the sensitivity of Orch-RM to teacher quality and orchestration format. Finally, in \appref{sec:appendix_rollouts}, we present qualitative examples of the orchestration rollouts of \ours and baselines across multiple domains of data.

\section{Limitations and Broader Impact}
\label{app:limitations}

\textbf{Limited scope of data and models.}\quad
Although \ours avoids expensive human annotations by leveraging existing MAS orchestration artifacts, its performance is still constrained by the diversity and quality of available orchestrator models and sampled trajectories. Due to the high computational cost of executing full MAS rollouts, we can only construct reward-model training data from subsets of the available domains and sampled trajectories, particularly for BrowseComp+, where trajectory diversity remains limited. Nevertheless, \ours still consistently outperforms competing selection and training methods under this restricted setting (\Tabref{tab:test-time-selection} and \Tabref{tab:training}). Developing larger-scale and more diverse orchestration datasets and benchmarks remains an important direction for future work.

\textbf{Separate reward models across domains.}\quad 
In this work, we train domain-specific reward models instead of a single unified reward model across all domains. While a unified reward model would improve scalability, different domains involve substantially different MAS orchestration patterns~\cite{ke2026masorchestraunderstandingimprovingmultiagent}. For instance, DeepScaleR and AIME 24\&25 primarily require minimal orchestration, whereas BrowseComp+ and HotpotQA involve medium-level orchestration. As shown in \Tabref{tab:ablation_rm_data}, directly combining these heterogeneous planning trajectories can introduce noise into the reward-model training data, making reward learning less stable and potentially reducing domain-specific accuracy. Developing a unified reward model that generalizes across diverse domains remains an important direction for future work.

{
\textbf{Dependence on teacher quality.}\quad
Our preference-construction pipeline relies on artifacts from existing orchestrators. As shown in \appref{app:teacher_quality}, reward models trained using artifacts from earlier MAS-Orchestra training steps (25--40) achieve lower Best@$N$ accuracy than those trained using the default teacher artifacts across all three evaluated benchmarks. These results indicate sensitivity to the quality of the available teacher artifacts and suggest that effectiveness may be limited when only weaker orchestrators are available. Reducing this dependence and extending the approach to settings without suitable pretrained orchestrators or execution logs remain important directions for future work.
}

\textbf{Broader Impact.}\quad 
This research advances methods for making multi-agent LLMs more capable and efficient through orchestration-level reward modeling, enabling earlier and more efficient intervention during MAS execution. Beyond improving performance, such orchestration-level supervision may also contribute to making MAS more controllable, trustworthy, and safe, as it allows system behaviors to be evaluated and guided before costly or potentially harmful sub-agent actions are fully executed. We hope this direction encourages future research on reliable oversight and governance mechanisms for increasingly autonomous multi-agent systems.

\section{Implementation Details}
\label{app:exp_settings}
We first present the implementation details of the post-training baselines in \appref{app:baseline_details}. Next, we describe the data specifications used for training \ours in \appref{app:dataset_details}. 
Finally, we include all prompt templates used throughout this paper in \appref{app:prompts}.

\subsection{Baselines}
\label{app:baseline_details}
We compare \ours against state-of-the-art post-training algorithms for MAS orchestrator training.

\begin{itemize}[nosep,leftmargin=24pt]
    \item \textbf{DPO}~\citep{Rafailov2023dpo}: We use the same paired dataset constructed for \ours and fine-tune the orchestration policy $\policy$ with the DPO objective for one epoch. 
    
    \item \textbf{RFT}~\citep{yuan2023rft}: Using the same \emph{``correct-over-incorrect''} data collected from MAS-Orchestra training logs, we perform rejection sampling by retaining only orchestrations that lead to correct final answers, followed by supervised fine-tuning of $\policy$. 
    
    \item \textbf{GRPO}~\citep{shao2024deepseekmath}: This baseline corresponds to continued training of MAS-Orchestra~\cite{ke2026masorchestraunderstandingimprovingmultiagent}, strictly following its original training configuration. 
    
    \item \textbf{LLM-as-a-Judge}~\citep{llm-as-a-judge}: We replace the original trajectory-level reward with scores generated by an LLM judge (\texttt{GPT-5-mini}), while keeping the remaining GRPO training setup unchanged.
\end{itemize}

\textbf{Shared Training Configuration.}\quad
Please refer to \Tabref{tab:config} for the shared training configuration.
\begin{table}[t]
    \centering
    \caption{Summary of training configuration.}
    \label{tab:dataset_details}
    \begin{tabular}{lc}
        \toprule
        \textbf{Setting} &  \textbf{Parameters} \\
        \midrule
        GPU  & 2 $\times$ RTX PRO 6000 96GB \\
        Batch size & 16 \\
        Actor rollout & 8 \\
        Clip ratio & 0.2 \\
        KL loss & 0.001 \\
        Learning rate & $1e{-}5$ \\
        Actor temperature & 0.7 \\
        Actor Top p & 0.9 \\
        LoRA Setting \\
        \quad rank $r$ & 32 \\
        \quad alpha $\alpha$ & 32 \\
        \quad target modules & All linear \\
        \bottomrule
    \end{tabular}
    \label{tab:config}
\end{table}



\subsection{Datasets}
\label{app:dataset_details}
\begin{table}[H]
    \centering
    \caption{Summary of dataset splits used for training, full inference, and testing. GPQA is evaluated strictly zero-shot as an out-of-distribution benchmark.}
    \label{tab:dataset_details}
    \begin{tabular}{llccc}
        \toprule
        \textbf{Dataset} & \textbf{Domain} & \textbf{Training} & \textbf{Full Inference} & \textbf{Testing} \\
        \midrule
        DeepScaleR / AIME & Mathematical Reasoning & 5,000 & 500 & 60 \\
        BrowseComp+ & Web-based QA & 1,066 & 500 & 200 \\
        HotpotQA & Multi-hop Reasoning & 5,000 & 2,000 & 200 \\
        GPQA & Scientific Reasoning (OOD) & -- & -- & 199 \\
        \bottomrule
    \end{tabular}
\end{table}

In this section, we provide a detailed breakdown of the datasets used across the different phases of our methodology: policy training, full inference (e.g., generating trajectories/rollouts for reward modeling), and final evaluation. Table~\ref{tab:dataset_details} summarizes the sample counts for each benchmark.

For mathematical reasoning, we utilize a subset of the DeepScaleR dataset for training and inference, while reserving the AIME 2024 and AIME 2025 problems strictly for testing. Similarly, GPQA is completely held out during the training and inference phases, serving exclusively as an out-of-distribution (OOD) evaluation to assess the cross-domain robustness of our trained orchestrator.

\vspace{1em}
\subsection{Prompt Templates}
\label{app:prompts}

\begin{tcolorbox}[
    breakable,
    enhanced,
    colback=blue!3!white,       
    colframe=blue!60!black,     
    boxrule=1pt,
    arc=3mm,
    left=5mm, right=5mm, top=4mm, bottom=4mm,
    title={\textbf{Prompt Template for DeepScaleR}}, 
    fonttitle=\bfseries\normalsize,
    coltitle=white,
    attach boxed title to top left={xshift=5mm, yshift=-3mm},
    boxed title style={colback=blue!60!black, arc=1.5mm}
]

\textbf{\texttt{<|start\_header\_id|>system<|end\_header\_id|>}}

\textbf{Cutting Knowledge Date:} December 2023\\
\textbf{Today Date:} 26 Jul 2024

\textbf{\texttt{<|eot\_id|><|start\_header\_id|>user<|end\_header\_id|>}}

\textbf{\texttt{<|im\_start|>system}}

You are a helpful assistant.

\textbf{MASness (How much Multi-Agent System-ness):} Minimal

\textbf{Valid Channels:} \texttt{thinking}, \texttt{agent}, \texttt{answer}

\textbf{Model:} \texttt{gpt-oss-120b}

An agent is a pre-configured AI personality that you can delegate tasks to. Each subagent:
\begin{enumerate}\itemsep0em
    \item Has a specific purpose and expertise area
    \item Uses its own context window separate from the main conversation
    \item (Optional) Can be configured with specific tools it's allowed to use
    \item Includes a custom system prompt that guides its behavior
\end{enumerate}

An agent should be defined in channel \textbf{\texttt{<agent>}}. Each agent must contain \textbf{\texttt{<agent\_name>}}, \textbf{\texttt{<agent\_description>}}, \textbf{\texttt{<required\_arguments>}}, and \textbf{\texttt{<agent\_output\_id>}}.

\textbf{DO NOT MISS ANY REQUEST FIELDS and ensure that your response is a well-formed XML object!}
\textbf{\texttt{<|im\_end|>}}

\vspace{1em}
\hrule
\vspace{1em}

\textbf{\texttt{<|im\_start|>user}}

Please solve the question step by step and create agents to delegate the task when necessary. First decide whether to solve directly or to delegate to exactly one agent. If you delegate, you must define that agent by outputting \textbf{\texttt{<agent>}} with \texttt{agent\_name} (select one of the agents: \texttt{CoTAgent, SCAgent, DebateAgent, ReflexionAgent}), \texttt{agent\_description}, \texttt{required\_arguments}, and \texttt{agent\_output\_id}. 

The final output of the delegated agent (identified by its \texttt{agent\_output\_id}) must represent the complete and final answer to the original question, not an intermediate result.

Always put either the final value or the \texttt{agent\_output\_id} in the \textbf{\texttt{<answer>}} tag, and use EXACTLY the same field names defined for the agent. If the selected agent uses roles (e.g., DebateAgent), also output \texttt{debate\_roles}.

\vspace{1em}
\textbf{For example:}

If you decide to solve the entire task yourself, you will output the following:

\begin{quote}
\textbf{Question:} What is (20+9)*(30+7)?

\textbf{\texttt{<thinking>}}\\
\quad The problem only requires basic arithmetic.\\
\quad No specialized reasoning agent or multi-agent discussion is needed.\\
\quad (20+9)*(30+7) = 600 + 140 + 270 + 63 = 1073.\\
\textbf{\texttt{</thinking>}}\\
\textbf{\texttt{<answer>}}1073\textbf{\texttt{</answer>}}
\end{quote}

If you decide to solve the entire task via delegation, you will output the following. In this case, since the \texttt{agent\_input} is the same as the original task, you must set the \texttt{agent\_input} as empty (""), and the parser will replace it with the original question.

\begin{quote}
\textbf{Question:} Compute the definite integral of (2x + 5) dx from 0 to 3.

\textbf{\texttt{<thinking>}}\\
\quad This problem involves symbolic integration and applying the Fundamental Theorem of Calculus.\\
\quad It requires structured reasoning rather than simple numeric computation.\\
\quad I will use a calculus agent that can perform step-by-step Chain-of-Thought reasoning.\\
\quad The final answer to the original question will be the output of the CoTAgent.\\
\textbf{\texttt{</thinking>}}\\
\textbf{\texttt{<agent>}}\\
\quad \textbf{\texttt{<agent\_name>}}CoTAgent\textbf{\texttt{</agent\_name>}}\\
\quad \textbf{\texttt{<agent\_description>}}Definite integrals with one Chain-of-Thought call.\textbf{\texttt{</agent\_description>}}\\
\quad \textbf{\texttt{<required\_arguments>}}\\
\quad \quad \textbf{\texttt{<agent\_input></agent\_input>}}\\
\quad \textbf{\texttt{</required\_arguments>}}\\
\quad \textbf{\texttt{<agent\_output\_id>}}calc\_agent\_output\textbf{\texttt{</agent\_output\_id>}}\\
\textbf{\texttt{</agent>}}\\
\textbf{\texttt{<answer>}}calc\_agent\_output\textbf{\texttt{</answer>}}
\end{quote}

\textit{[Additional examples formatted identically...]}

\vspace{1em}
\textbf{Below is the question to solve:}

$N$ students are seated at desks in an $m \times n$ array, where $m, n \ge 3$. Each student shakes hands with the students who are adjacent horizontally, vertically or diagonally. If there are 1020 handshakes, what is $N$?

\textbf{\texttt{<|im\_end|>}}

\vspace{1em}
\hrule
\vspace{1em}

\textbf{\texttt{<|im\_start|>assistant}}

\textbf{Channels:}
\begin{itemize}\itemsep0em
    \item \textbf{\texttt{<thinking>}}: internal reasoning and planning
    \item \textbf{\texttt{<agent>}}: definition of agents
    \item \textbf{\texttt{<answer>}}: final user-facing answer
\end{itemize}

\textbf{Models (the Large Language Model used in agent):}
\begin{itemize}\itemsep0em
    \item \textbf{gpt-4.1-nano:} Fastest, most cost-efficient version. Excels at instruction following and tool calling. Features a 1M token context window, low latency without a reasoning step. Jun 01, 2024 knowledge cutoff.
    \item \textbf{gpt-oss-120b:} State-of-the-art language model delivering strong real-world performance at low cost. Outperforms similarly sized open models on reasoning tasks, demonstrates strong tool use, and is optimized for efficient deployment. 131,072 context window, Jun 01, 2024 knowledge cutoff, Reasoning token support.
\end{itemize}

\textbf{MASness Levels:}
\begin{itemize}\itemsep0em
    \item \textbf{Minimal:} direct solve or at most one agent
    \item \textbf{Medium:} one or more agents delegation
    \item \textbf{High:} complex multi-agent delegation
\end{itemize}

\textbf{Sub-agent Schema (all fields required):}
\begin{verbatim}
<agent>
    <agent_name>...</agent_name> (CoTAgent, SCAgent, 
    
    DebateAgent, ReflexionAgent)
    <agent_description>...</agent_description>
    <required_arguments>
        <...>...</...>
    </required_arguments>
    <agent_output_id>...</agent_output_id>
</agent>
\end{verbatim}

\vspace{1em}
\textbf{Available Agents \& Implementations:}

\vspace{0.5em}
\textbf{1. Chain-of-Thought Agent (CoTAgent)}
\begin{description}
    \item[Description:] Encourages the LLM to think step by step rather than directly outputting an answer, enabling complex problem-solving.
    \item[Required Arguments:] \texttt{agent\_input} (If left empty, defaults to original question).
\end{description}
\begin{quote}\ttfamily\scriptsize\color{blue!20!black}
async def CoTAgent(self, agent\_input, model: str):\\
\hspace*{4mm}from mas\_r1\_reasoner.agents.agent\_system import LLMAgentBase, Info\\
\hspace*{4mm}assert isinstance(agent\_input, Info), f"agent\_input must be an Info object"\\
\hspace*{4mm}temperature = 0.5\\
\hspace*{4mm}cot\_instruction = "Please think step by step and then solve the task."\\
\hspace*{4mm}cot\_agent = LLMAgentBase(['thinking', 'answer'], 'Chain-of-Thought LLM', model, temperature)\\
\hspace*{4mm}thinking, answer = await cot\_agent([agent\_input], cot\_instruction)\\
\hspace*{4mm}return self.make\_final\_answer(thinking, answer)
\end{quote}

\vspace{0.5em}
\textbf{2. Self-Consistency with Chain-of-Thought (SCAgent)}
\begin{description}
    \item[Description:] Repeatedly asks the same question with high temperature to generate different reasoning paths, combining them through majority voting ensembling.
    \item[Required Arguments:] \texttt{agent\_input}.
\end{description}
\begin{quote}\ttfamily\scriptsize\color{blue!20!black}
async def SCAgent(self, agent\_input, model: str):\\
\hspace*{4mm}from mas\_r1\_reasoner.agents.agent\_system import LLMAgentBase, Info\\
\hspace*{4mm}temperature = 0.5; num\_repeated\_samples = 5\\
\hspace*{4mm}cot\_instruction =  "Please think step by step and then solve the task."\\
\hspace*{4mm}cot\_agents = [LLMAgentBase(['thinking', 'answer'], 'Chain-of-Thought LLM', model, temperature) for \_ in range(num\_repeated\_samples)]\\
\hspace*{4mm}possible\_answers = []\\
\hspace*{4mm}for i in range(num\_repeated\_samples):\\
\hspace*{8mm}thinking, answer = await cot\_agents[i]([agent\_input], cot\_instruction)\\
\hspace*{8mm}possible\_answers.append(answer.content)\\
\hspace*{4mm}answer = self.majority\_voting(possible\_answers)\\
\hspace*{4mm}return self.make\_final\_answer(thinking, answer)
\end{quote}

\vspace{0.5em}
\textbf{3. LLM Debate (DebateAgent)}
\begin{description}
    \item[Description:] Lets different LLMs debate with each other, leveraging diverse perspectives to find better solutions.
    \item[Required Arguments:] \texttt{agent\_input}, \texttt{debate\_roles} (List of roles, e.g., ["Math Professor", "Statistician"]).
\end{description}
\begin{quote}\ttfamily\scriptsize\color{blue!20!black}
async def DebateAgent(self, agent\_input, model: str, debate\_roles: List[str]):\\
\hspace*{4mm}temperature = 0.5; max\_debate\_round = 5\\
\hspace*{4mm}debate\_agents = [LLMAgentBase(..., role=role) for role in debate\_roles]\\
\hspace*{4mm}for r in range(max\_debate\_round):\\
\hspace*{8mm}for i in range(len(debate\_agents)):\\
\hspace*{12mm}\# Agents review each other's previous thoughts and update\\
\hspace*{12mm}thinking, answer = await debate\_agents[i](input\_infos, debate\_instruction)\\
\hspace*{4mm}thinking, answer = await final\_decision\_agent(..., final\_decision\_instruction)\\
\hspace*{4mm}return self.make\_final\_answer(thinking, answer)
\end{quote}

\vspace{0.5em}
\textbf{4. Self-Refine (ReflexionAgent)}
\begin{description}
    \item[Description:] Iteratively improves answers based on feedback from a Critic agent.
    \item[Required Arguments:] \texttt{agent\_input}.
\end{description}
\begin{quote}\ttfamily\scriptsize\color{blue!20!black}
async def ReflexionAgent(self, agent\_input, model: str):\\
\hspace*{4mm}temperature = 0.5; max\_reflection\_round = 5\\
\hspace*{4mm}cot\_agent = LLMAgentBase(['thinking', 'answer'], 'Chain-of-Thought LLM', ...)\\
\hspace*{4mm}critic\_agent = LLMAgentBase(['feedback', 'correct'], 'Critic LLM', ...)\\
\hspace*{4mm}thinking, answer = await cot\_agent(cot\_inputs, initial\_instruction, 0)\\
\hspace*{4mm}for i in range(max\_reflection\_round):\\
\hspace*{8mm}feedback, correct = await critic\_agent([agent\_input, thinking, answer], ...)\\
\hspace*{8mm}if correct.content == 'True': break\\
\hspace*{8mm}thinking, answer = await cot\_agent(..., reflect\_instruction, i + 1)\\
\hspace*{4mm}return self.make\_final\_answer(thinking, answer)
\end{quote}

\vspace{1em}
\textbf{\texttt{<|im\_end|><|eot\_id|><|start\_header\_id|>assistant<|end\_header\_id|>}}

\textbf{\texttt{<thinking>}}\\
\quad To solve this problem, we need to determine the number of handshakes that occur in an $m \times n$ grid where each student shakes hands with their adjacent neighbors (horizontally, vertically, and diagonally). The total number of handshakes is given as 1020. We need to find the number of students $N$ in the grid.\\
\textbf{\texttt{</thinking>}}

\textbf{\texttt{<agent>}}\\
\quad \textbf{\texttt{<agent\_name>}}DebateAgent\textbf{\texttt{</agent\_name>}}\\
\quad \textbf{\texttt{<agent\_description>}}Near-tie numeric choice using one Debate call.\textbf{\texttt{</agent\_description>}}\\
\quad \textbf{\texttt{<required\_arguments>}}\\
\quad \quad \textbf{\texttt{<agent\_input></agent\_input>}}\\
\quad \quad \textbf{\texttt{<debate\_roles>}}["Graph Theory Expert", "Combinatorics Specialist"]\textbf{\texttt{</debate\_roles>}}\\
\quad \textbf{\texttt{</required\_arguments>}}\\
\quad \textbf{\texttt{<agent\_output\_id>}}debate\_agent\_output\textbf{\texttt{</agent\_output\_id>}}\\
\textbf{\texttt{</agent>}}\\
\textbf{\texttt{<answer>}}debate\_agent\_output\textbf{\texttt{</answer>}}

In this problem, the number of handshakes can be complex to calculate directly without a structured approach. The DebateAgent will help us consider different combinatorial and graph theory perspectives to ensure we reach the correct solution.
\textbf{\texttt{<|eot\_id|>}}

\end{tcolorbox}

\section{Additional Analyses}
\label{app:additional_analyses}
\subsection{Reliability of Constructed Preferences}
\label{app:preference_reliability}
We construct SoB preferences by treating orchestration plans generated by the domain-specialized orchestrator as preferred over those generated by the base model for the same input. For CoI preferences, we compare plans for the same input and treat those whose executions produce correct final answers as preferred over those whose executions produce incorrect answers. Neither guarantees that the preferred plan will outperform the other plan in every execution, since sub-agent execution is stochastic. We therefore evaluate the reliability of both preference sources through repeated executions of fixed orchestration plans.

\paragraph{Experimental setup.}
We sample SoB and CoI preference pairs from the DeepScaleR, BrowseComp+, and HotpotQA training datasets. For each pair, we fix both orchestration plans and execute each plan five times using different sampling seeds. We estimate each plan's success probability by the proportion of executions producing a correct final answer, and compare these estimates to assess agreement with the original preference label.

\begin{table}[h]
    \centering
    \caption{Pairwise win rates of the originally preferred orchestration plans under repeated execution. Each fixed plan is executed five times. Values are percentages.}
    \label{tab:preference_reliability}
    \begin{tabular}{lcc}
        \toprule
        Dataset & CoI & SoB \\
        \midrule
        DeepScaleR  & 68.00 & 94.00 \\
        BrowseComp+ & 67.50 & 96.50 \\
        HotpotQA    & 72.50 & 96.00 \\
        \bottomrule
    \end{tabular}
\end{table}

\paragraph{Results.}
As shown in Table~\ref{tab:preference_reliability},
SoB preferences achieve pairwise win rates of
94.0--96.5\%, whereas CoI preferences achieve
67.5--72.5\%.
Both sources therefore retain useful preference information under repeated execution, although agreement is lower for CoI. These results support their use as supervision while highlighting the potential benefit of constructing preferences from aggregated execution outcomes.

\subsection{Orchestration Diversity Across Benchmarks}
\label{app:orchestration_diversity}

To investigate the uneven benefits of orchestration-level reward modeling across benchmarks, we analyze generated orchestration plans along two dimensions:
(i) semantic diversity, based on cosine similarity between plan embeddings, and
(ii) structural diversity, characterized by the number and types of selected sub-agents.
Table~\ref{tab:orchestration_diversity} summarizes the results.

\begin{table}[H]
    \centering
    \caption{Orchestration diversity across datasets. Sub-agent-count distributions and agent-type shares are separately normalized to 100\% before rounding. Values are reported to two decimal places; displayed totals may differ slightly from 100\%.}
    \label{tab:orchestration_diversity}
    \small
    \setlength{\tabcolsep}{3pt}

    \textbf{Semantic diversity and sub-agent counts}\\[3pt]
    \begin{tabular}{lrrrrrrrr}
        \toprule
        Dataset & \shortstack{Semantic\\diversity}
        & 1 & 2 & 3 & 4 & 5 & 6 & $>6$ \\
        \midrule
        DeepScaleR
        & 0.12 & 86.65 & 9.04 & 1.63 & 1.20
        & 0.47 & 0.33 & 0.67 \\
        BrowseComp+
        & 0.15 & 0.02 & 0.04 & 1.71 & 37.48
        & 45.21 & 9.87 & 5.67 \\
        HotpotQA
        & 0.04 & 0.00 & 99.89 & 0.11 & 0.00
        & 0.00 & 0.00 & 0.00 \\
        GPQA
        & 0.19 & 99.75 & 0.25 & 0.00 & 0.00
        & 0.00 & 0.00 & 0.00 \\
        \bottomrule
    \end{tabular}

    \vspace{8pt}
    \textbf{Agent-type composition}\\[3pt]
    \begin{tabular}{lrrrrr}
        \toprule
        Dataset & CoT & Debate & Reflexion & Web Search & SC \\
        \midrule
        DeepScaleR
        & 0.00 & 55.45 & 44.52 & 0.00 & 0.02 \\
        BrowseComp+
        & 15.48 & 0.00 & 17.69 & 66.81 & 0.02 \\
        HotpotQA
        & 50.00 & 0.00 & 0.00 & 50.00 & 0.00 \\
        GPQA
        & 0.00 & 83.17 & 16.83 & 0.00 & 0.00 \\
        \bottomrule
    \end{tabular}
\end{table}

\paragraph{Results.}
HotpotQA has the lowest reported semantic-diversity score, and 99.89\% of its plans use exactly two sub-agents. Its agent-type composition is evenly divided between CoT and Web Search. GPQA is also structurally homogeneous: 99.75\% of its plans use a single sub-agent, with Debate and Reflexion accounting for 83.17\% and 16.83\% of the agent-type composition, respectively. Importantly, GPQA does not exhibit low semantic diversity under the reported metric and its homogeneity is primarily structural. These patterns suggest that limited structural variation may leave fewer opportunities for reward-guided plan selection, potentially contributing to the smaller gains on these benchmarks.

\subsection{Computational Cost of Continued Orchestrator Training}
\label{app:training_cost}

We compare the estimated computational costs of GRPO and \ours-guided continued orchestrator training. The comparison includes upfront reward-model training, orchestrator sampling and optimization, and reward computation.

\paragraph{Cost estimation.} 
Let $C$ denote computational cost in FLOPs, and let $P_o$, $P_r$, and $P_s$ denote the parameter counts of the orchestrator, reward model, and sub-agent model, respectively. In our experiments, we use a 7B orchestrator ($P_o = 7 \times 10^9$), an 8B reward model ($P_r = 8 \times 10^9$), and a 120B sub-agent model ($P_s = 120 \times 10^9$). We approximate training and inference costs by $6PT$ and $2PT$, respectively, where $T$ denotes the number of
processed tokens.

Our continued-training experiments use $N=8$ parallel candidate plans per query for both GRPO and \ours. The matched computational-cost comparison uses this same configuration. We additionally report estimated GRPO costs at $N=32$ as a separate reference corresponding to its default configuration.

The cost components are defined as follows:

\begin{enumerate}
    \item \textbf{Reward-model training.}
    Let $C_r^{\mathrm{train}}$ denote the total cost of \ours training, and $T_r^{\mathrm{train}}$ denote the total training tokens processed by \ours.
    The upfront training cost is $C_r^{\mathrm{train}} \approx 6P_r T_r^{\mathrm{train}}$.
    This component applies only to \ours.

    \item \textbf{Orchestrator inference.}
    Let $C_o^{\mathrm{infer}}$ denote the total cost of orchestrator inference, and $T_o^{\mathrm{infer}}$ denote the total inference tokens processed when generating candidate orchestration plans.
    The inference cost is $C_o^{\mathrm{infer}} \approx 2P_o T_o^{\mathrm{infer}}$.
    This component applies to both GRPO and \ours.

    \item \textbf{Orchestrator update.}
    Let $C_o^{\mathrm{train}}$ denote the total cost of the orchestrator update, and $T_o^{\mathrm{train}}$ denote the total training tokens processed during orchestrator optimization.
    The update cost is $C_o^{\mathrm{train}} \approx 6P_o T_o^{\mathrm{train}}$.
    This component applies to both GRPO and \ours.

    \item \textbf{Sub-agent rollouts.}
    Let $C_s^{\mathrm{infer}}$ denote the total cost of sub-agent execution, and $T_s^{\mathrm{infer}}$ denote the total inference tokens processed by sub-agents when executing candidate plans.
    For GRPO, the reward-computation cost is $C_s^{\mathrm{infer}} \approx 2P_s T_s^{\mathrm{infer}}$.
    This component applies only to GRPO.

    \item \textbf{Reward-model scoring.}
    Let $C_r^{\mathrm{infer}}$ denote the total cost of \ours inference on the orchestration plans, and $T_r^{\mathrm{infer}}$ denote the total inference tokens processed by the reward model when scoring candidate plans.
    For \ours, the reward-computation cost is $C_r^{\mathrm{infer}} \approx 2P_r T_r^{\mathrm{infer}}$.
    This component applies only to \ours.
\end{enumerate}

Token counts are accumulated separately for each method and component over the full sampling and training process. The total GRPO cost is $C_o^{\mathrm{infer}} + C_o^{\mathrm{train}} + C_s^{\mathrm{infer}}$, whereas the total \ours cost is
$C_r^{\mathrm{train}} + C_o^{\mathrm{infer}} + C_o^{\mathrm{train}} + C_r^{\mathrm{infer}}$.

All costs are reported in exaFLOPs (EF), where $1\,\mathrm{EF}=10^{18}$ FLOPs. These are analytical estimates of computational cost, rather than measured wall-clock runtimes.

\begin{table}[H]
    \centering
    \caption{Estimated computational costs of continued orchestrator training, in EF. $N$ denotes the number of parallel candidate samples. Total costs include upfront reward-model training for \ours.}
    \label{tab:training_cost_breakdown}
    \small
    \setlength{\tabcolsep}{3pt}
    \begin{tabular}{lrrrrrrr}
        \toprule
        Method & $N$
        & \shortstack{\ours\\training}
        & \shortstack{Orch.\\inference}
        & \shortstack{Orch.\\update}
        & \shortstack{Sub-agent\\rollouts}
        & \shortstack{\ours\\scoring}
        & Total \\
        \midrule

        \multicolumn{8}{l}{\textbf{DeepScaleR}} \\
        \midrule
        GRPO (default)
        & 32 & 0.00 & 18.84 & 2.68 & 321.44 & 0.00
        & 342.96 \\
        GRPO
        & 8 & 0.00 & 4.71 & 0.67 & 80.36 & 0.00
        & 85.74 \\
        \ours
        & 8 & 1.06 & 4.71 & 0.67 & 0.00 & 2.82
        & \textbf{9.26} \\
        \midrule

        \multicolumn{8}{l}{\textbf{BrowseComp+}} \\
        \midrule
        GRPO (default)
        & 32 & 0.00 & 7.24 & 1.32 & 1028.44 & 0.00
        & 1037.00 \\
        GRPO
        & 8 & 0.00 & 1.81 & 0.33 & 257.11 & 0.00
        & 259.25 \\
        \ours
        & 8 & 1.09 & 1.81 & 0.33 & 0.00 & 0.78
        & \textbf{4.01} \\
        \midrule

        \multicolumn{8}{l}{\textbf{HotpotQA}} \\
        \midrule
        GRPO (default)
        & 32 & 0.00 & 28.80 & 5.80 & 130.88 & 0.00
        & 165.48 \\
        GRPO
        & 8 & 0.00 & 7.20 & 1.45 & 32.72 & 0.00
        & 41.37 \\
        \ours
        & 8 & 1.34 & 7.20 & 1.45 & 0.00 & 3.11
        & \textbf{13.10} \\
        \bottomrule
    \end{tabular}
\end{table}

\paragraph{Results.}
Under the matched $N=8$ setting,
\Tabref{tab:training_cost_breakdown} shows that GRPO requires approximately 9.3, 64.7, and 3.2 times the estimated FLOPs of \ours on DeepScaleR, BrowseComp+, and HotpotQA, respectively.
These reductions persist after including the upfront cost of reward-model training. The largest reduction occurs on BrowseComp+, where sub-agent rollouts dominate the GRPO cost. These results highlight the computational benefit of replacing full sub-agent execution with orchestration-level reward scoring during continued training.

\subsection{Sensitivity to Teacher Quality}
\label{app:teacher_quality}

We investigate how the quality of the source orchestrator affects the resulting orchestration reward model. We train a weaker-teacher variant, denoted as \ours (WT), using artifacts from early MAS-Orchestra training steps (25--40), following the same preference-construction procedure as in our main setting. The default model instead uses artifacts from the domain-specialized teachers. \Tabref{tab:teacher_statistics} summarizes the teacher checkpoints and their average accuracy.

\begin{table}[H]
    \centering
    \caption{Comparison of weaker and default teachers. WT denotes the weaker teacher. Steps refer to MAS-Orchestra training steps. Accuracy values are percentages.}
    \label{tab:teacher_statistics}
    \begin{tabular}{lrrrr}
        \toprule
        Benchmark
        & \shortstack{WT\\steps}
        & \shortstack{WT\\avg. acc.}
        & \shortstack{Default\\step}
        & \shortstack{Default\\avg. acc.} \\
        \midrule
        AIME        & 25--40 & 48.75 & 180 & 58.96 \\
        BrowseComp+ & 25--40 &  6.25 & 140 &  9.44 \\
        HotpotQA    & 25--40 & 40.25 & 250 & 40.63 \\
        \bottomrule
    \end{tabular}
\end{table}

We then compare the Best-of-$N$ selection performance of \ours (WT) and the default \ours.
Oracle selection is included as a reference.

\begin{table}[H]
    \centering
    \caption{Best-of-$N$ accuracy of reward models trained using weaker and default teacher artifacts. Values are percentages.}
    \label{tab:teacher_quality_results}
    \begin{tabular}{lrrr}
        \toprule
        Benchmark & Oracle & \ours (WT) & \ours \\
        \midrule
        AIME        & 76.67 & 40.00 & \textbf{68.33} \\
        BrowseComp+ & 24.50 &  9.00 & \textbf{14.00} \\
        HotpotQA    & 57.00 & 36.50 & \textbf{42.50} \\
        \bottomrule
    \end{tabular}
\end{table}

\paragraph{Results.}
As shown in \Tabref{tab:teacher_quality_results}, the default \ours outperforms the weaker-teacher variant across all three evaluated benchmarks. These results indicate that the choice of teacher artifacts affects reward-guided orchestration selection, with artifacts from the default teachers yielding stronger performance than those from earlier training stages.

\subsection{Sensitivity to Orchestration Format}
\label{app:format_transfer}

We investigate whether \ours can score orchestration plans expressed in a format different from its training representation.

\paragraph{Experiment setup.}We use GPT-5-mini to generate natural-language (NL) orchestration plans and convert the same plans into XML. We then score both representations using \ours and an LLM-as-a-Judge baseline. Sub-agent execution uses the XML representation in both conditions, allowing us to compare selection performance when the scoring input is expressed in NL or XML.

\begin{table}[H]
    \centering
    \caption{Orchestration selection performance with natural-language (NL) and XML scoring inputs. Both representations correspond to the same generated plans, and sub-agent execution uses XML. Accuracy values are percentages.}
    \label{tab:format_transfer_results}
    \small
    \setlength{\tabcolsep}{4pt}
    \begin{tabular}{lrrrrrr}
        \toprule
        Benchmark
        & \shortstack{Average\\acc.}
        & Oracle
        & \multicolumn{2}{c}{\ours Best@$N$}
        & \multicolumn{2}{c}{LLM Judge} \\
        \cmidrule(lr){4-5}
        \cmidrule(lr){6-7}
        & & & NL & XML & NL & XML \\
        \midrule
        AIME
        & 49.33 & 76.67 & 50.00 & 56.66 & 50.00 & 50.00 \\
        BrowseComp+
        & 3.50 & 13.50 & 3.50 & 7.00 & 3.50 & 5.50 \\
        HotpotQA
        & 35.44 & 59.50 & 34.00 & 38.50 & 35.00 & 38.50 \\
        \bottomrule
    \end{tabular}
\end{table}

\paragraph{Results.}
As shown in \Tabref{tab:format_transfer_results}, \ours achieves higher selection accuracy with XML inputs than with NL inputs across all three benchmarks. Direct NL scoring provides little or no improvement over average candidate accuracy, whereas XML scoring improves selection performance. Even when the plans are generated by GPT-5-mini rather than MAS-Orchestra, \ours matches or exceeds the LLM-as-a-Judge baseline across all three benchmarks when scoring their XML representations.

These results suggest that the transferability of \ours is currently format-sensitive, but its learned preference signal remains effective for plan selection when new orchestration plans are mapped into the XML representation used during training.

\section{Orchestration Rollout Examples: Base vs. Trained Model}
\label{sec:appendix_rollouts}

To empirically demonstrate the impact of our Orchestration Reward Modeling, we provide raw rollout samples from the AIME, BrowseComp+ and HotpotQA benchmarks. In the following sections, we define our two comparative stages as follows:
\begin{itemize}
    \item \textbf{Base Model:} The baseline checkpoint trained entirely through the original MAS-Orchestra pipeline.
    \item \textbf{Trained Model:} The resulting policy after continual training from the baseline checkpoint using our proposed reward modeling framework.
\end{itemize}

These samples highlight a massive distributional shift in policy behavior, demonstrating how our continual training resolves the structural and planning bottlenecks inherent in the baseline framework. Poor orchestration plans directly result in subagent failure, while strict, mathematically sound plans yield correct ground truth answers.
 \subsection{Base Model Rollouts}
\label{sec:appendix_base_rollouts}
\begin{promptbox}[Task: AIME 2025 (ID: aime25\_11)]
The set of points in 3-dimensional coordinate space that lie in the plane $x+y+z=75$ whose coordinates satisfy the inequalities $x-yz<y-zx<z-xy$ forms three disjoint convex regions. Exactly one of those regions has finite area. The area of this finite region can be expressed in the form $a\sqrt{b}$, where $a$ and $b$ are positive integers and $b$ is not divisible by the square of any prime. Find $a+b$.
\end{promptbox}
\begin{basebox}[Base Model: Sample 1 ]]
<thinking>
    This problem involves geometric and algebraic concepts. The set of points in 3-dimensional space that lie in the plane \(x + y + z = 75\) and satisfy the inequalities \(x - yz \textless y - zx \textless z - xy\) can be analyzed by breaking it down into smaller, manageable parts. The key is to understand the geometric regions defined by these inequalities and their intersection with the plane.
</thinking>
<agent>
    <agent_name>DebateAgent</agent_name>
    <agent_description>Near-tie numeric choice using one Debate call.</agent_description>
    <required_arguments>
        <agent_input></agent_input>
        <debate_roles>["Geometric Analyst", "Algebraic Solver"]</debate_roles>
    </required_arguments>
    <agent_output_id>
    debate_agent_output
    </agent_output_id>
</agent>
<answer>debate_agent_output</answer>

Here, we have delegated the problem to the DebateAgent, which will handle the detailed geometric and algebraic analysis. The two roles, "Geometric Analyst" and "Algebraic Solver," will debate and to find the final the final the inequalities and the intersection with the plane, ultimately providing the correct region.

----------------------------------------------------------------------
> Executing DebateAgent [Geometric Analyst vs. Algebraic Solver]...
> Analyst: Projection onto xy-plane yields a triangle of area 507/4. 
> Solver: Derived side lengths based on plane x+y+z=75, area is 150*sqrt(3).
> Resulting Rationale: Conflict in coordinate derivation led to summation error.
> Final Answer Produced: 1017
----------------------------------------------------------------------
Result: [Incorrect] (Ground Truth: 510)
\end{basebox}

\vspace{1em}

\begin{basebox}[Base Model: Sample 4 ]
<thinking>
    This problem involves geometric and algebraic reasoning, specifically dealing with the intersection of a plane and inequalities. The key steps involve understanding the plane equation and the inequalities, which can be complex. I will delegate this to the ReflexionAgent to ensure accuracy and thoroughness.
</thinking>
<agent>
    <agent_name>ReflexionAgent</agent_name>
    <agent_description>Solves reasoning tasks with iterative self-refine loop (single agent call).</agent_description>
    <required_arguments>
        <agent_input></agent_input>
    </required_arguments>
    <agent_output_id>
    reflexion_agent_output
    </agent_output_id>
</agent>
<answer>reflexion_agent_output</answer>

The ReflexionAgent will handle the detailed reasoning and ensure that the solution is accurate and complete. The ReflexionAgent will provide the final answer, which will be in the form \(a\sqrt{b}\).

----------------------------------------------------------------------
> Executing ReflexionAgent...
> Subagent thinking: Triangle vertices at (0,0,75), (0,75,0), (75,0,0)...
> Side length is 75, area is 2812.5*sqrt(3).
> Final Answer Produced: 903
----------------------------------------------------------------------
Result: [Incorrect] (Ground Truth: 510)
\end{basebox}
\vspace{1em}

\begin{basebox}[Base Model: Sample 7]
<thinking>
    To solve this problem, we need to identify the regions in the plane \(x + y + z = 75\) that satisfy the inequalities \(x - yz\textless y - zx\textless z - xy\). Lets analyze the inequalities step by step.
</thinking>
<agent>
    <agent_name>DebateAgent</agent_name>
    <agent_description>Near-tie numeric choice using one Debate call.</agent_description>
    <required_arguments>
        <agent_input></agent_input>
        <debate_roles>["Mathematics Professor", "Geometry Specialist"]</debate_roles>
    </required_arguments>
    <agent_output_id>
    debate_agent_output
    </agent_output_id>
</agent>
<answer>debate_agent_output</answer>

### Detailed Explanation of the Agent Delegation

1. Initial Analysis:
   - The plane equation \(x + y + z = 75\) represents a plane in 3D space.
   - The inequalities \(x - yz\textless y - zx\textless z - xy\) need to be analyzed to determine the regions where they hold true.

2. Agent into the De De of the inequalities:
   - \(x - yz\textless y - zx\)
   - \(y - zx\textless z - xy\)

3. Delegation to DebateAgent:
   - We need to debate the regions where these inequalities hold true.
   - The DebateAgent will involve perspectives from a "Mathematics Professor" and a "Geometry Specialist" to ensure a comprehensive and accurate solution.

### Agent Delegation to DebateAgent

The DebateAgent will handle the detailed analysis and provide a final answer based on the perspectives of the two roles.

<thinking>
    To solve this problem, we need to identify the regions in the plane \(x + y + z = 75\) that satisfy the inequalities \(x - yz\textless y - zx\textless z - xy\). Lets analyze the inequalities step by step.
</thinking>
<agent>
    <agent_name>DebateAgent</agent_name>
    <agent_description>Near-tie numeric choice using one Debate call.</agent_description>
    <required_arguments>
        <agent_input></agent_input>
        <debate_roles>["Mathematics Professor", "Geometry Specialist"]</debate_roles>
    </required_arguments>
    <agent_output_id>
    debate_agent_output
    </agent_output_id>
</agent>
<answer>debate_agent_output</answer>

----------------------------------------------------------------------
> Executing DebateAgent...
> System Error: Redundant XML blocks detected. 
> Subagent thinking: Intersection yields region bounded by x=-1, y=-1, z=-1.
> Area calculation resulted in triangle with area 300*sqrt(3).
> Final Answer Produced: 303
----------------------------------------------------------------------
Result: [Incorrect] (Ground Truth: 510)
\end{basebox}

\vspace{1em}

\subsection{Our Trained Orchestrator Rollouts}
\label{sec:trained_planning_rollout}

\begin{promptbox}[Task: AIME 2025 (ID: aime25\_11)]
The set of points in 3-dimensional coordinate space that lie in the plane $x+y+z=75$ whose coordinates satisfy the inequalities $x-yz<y-zx<z-xy$ forms three disjoint convex regions. Exactly one of those regions has finite area. The area of this finite region can be expressed in the form $a\sqrt{b}$, where $a$ and $b$ are positive integers and $b$ is not divisible by the square of any prime. Find $a+b$.
\end{promptbox}

\vspace{1em}

\begin{trainedbox}[Trained Model: Planning Response (Sample 1)]
<thinking>
    Compute the area of the finite convex region defined by $x+y+z=75$ and $x-yz\textless y-zx\textless z-xy$.
</thinking>
<agent>
    <agent_name>DebateAgent</agent_name>
    <agent_description>Near-tie numeric choice and geometric/inequality constraints resolution using one Debate call.</agent_description>
    <required_arguments>
        <agent_input></agent_input>
        <debate_roles>["Geometry Specialist", "Inequality Expert"]</debate_roles>
    </required_arguments>
    <agent_output_id>
    debate_agent_output
    </agent_output_id>
</agent>
<answer>debate_agent_output</answer>

Explanation:

1. Thinking Phase:
   - We comprehend the given problem: Finding the area of a finite convex region in 3D space defined by the plane equation \(x + y + z = 75\) and the inequalities \(x - yz\textless y - zx\textless z - xy\).

2. Delegation Decision:
   - This problem combines geometry (plane and convex regions), algebra (inequalities), and area computation. It benefits from diverse perspectives, making it ideal for the DebateAgent. The DebateAgent allows different specialized LLMs (e.g., a "Geometry Specialist" and an "Inequality Expert") to debate and converge on the correct solution.
   - DebateAgent is chosen over others (like CoTAgent, SCAgent, or ReflexionAgent) because:
     - CoTAgent focuses on step-by-step reasoning but doesn't leverage multiple viewpoints.
     - SCAgent aims for consensus through repeated sampling, which is overkill here.
     - ReflexionAgent is better suited for self-correction after initial attempts, not initial complex problem-solving.

3. Agent Definition:
   - Agent Name: DebateAgent
   - Agent Description: Fits the problem's need for diverse and justified reasoning.
   - Required Arguments:
     - agent_input: Set to "" (empty) as instructed, ensuring the original question is inserted by the parser.
     - debate_roles: A list with two distinct expert roles ("Geometry Specialist" and "Inequality Expert"), promoting a rich debate.
   - Agent Output ID: debate_agent_output to reference the final answer from the delegated agent.

4. Final Answer Tag:
   - The <answer> tag will contain the output identifier debate_agent_output, representing the complete and verified solution generated by the DebateAgent.

This approach ensures both correctness and confidence through collaborative reasoning.

----------------------------------------------------------------------
> Executing DebateAgent [Geometry Specialist vs. Inequality Expert]...
> Specialist: Identified boundary lines x=y, y=z, and z=-1 on the plane.
> Expert: Verified bounded region occurs when x > -1, z > -1, forcing x < y < z.
> Consensus: Triangle area calculated via cross product yields 507*sqrt(3).
> Final Answer Produced: 510
----------------------------------------------------------------------
Result: [Correct] (Ground Truth: 510)
\end{trainedbox}

\vspace{1em}

\begin{trainedbox}[Trained Model: Planning Response (Sample 4)]
<thinking>
    Compute the area of the finite region defined by $x+y+z=75$ and $x-yz \textless y-zx \textless z-xy$.
</thinking>
<agent>
    <agent_name>DebateAgent</agent_name>
    <agent_description>Near-tie numeric choice and geometric/inequality constraint satisfaction using one Debate call.</agent_description>
    <required_arguments>
        <agent_input></agent_input>
        <debate_roles>["Geometry Specialist", "Inequality Expert"]</debate_roles>
    </required_arguments>
    <agent_output_id>
    debate_agent_output
    </agent_output_id>
</agent>
<answer>debate_agent_output</answer>

Explanation:

1. Thinking Phase:
   - We understand the given constraints: the plane equation \(x + y + z = 75\) and the inequalities \(x - yz\textless y - zx\textless z - xy\). The goal is to find the finite-area convex region formed by these conditions and express its area in the form \(a\sqrt{b}\), then compute \(a + b\).

2. Delegation Decision:
   - This problem combines geometry (plane equations and region definitions), algebra (inequalities), and area computation. It benefits from diverse perspectives, making it ideal for the DebateAgent. The DebateAgent allows different specialized LLMs (e.g., a "Geometry Specialist" and an "Inequality Expert") to debate and converge on the correct solution. This agent is chosen over others (like CoTAgent, SCAgent, or ReflexionAgent) because:
     - CoTAgent focuses on step-by-step reasoning but doesn't leverage multiple viewpoints.
     - SCAgent aims for consensus through repeated sampling, which is overkill here.
     - ReflexionAgent is better suited for self-correction after initial attempts, not initial complex problem-solving.

3. Agent Definition:
   - Agent Name: DebateAgent
   - Agent Description: Fits the problem's need for diverse and justified reasoning.
   - Required Arguments:
     - agent_input: Set to "" (empty) as instructed, ensuring the original question is inserted by the parser.
     - debate_roles: A list containing two distinct expert roles ("Geometry Specialist" and "Inequality Expert"), promoting a rich debate.
   - Agent Output ID: debate_agent_output to reference the final answer from the delegated agent.

4. Final Answer Tag:
   - The <answer> tag will contain the output identifier debate_agent_output, representing the complete and justified solution generated by the DebateAgent.

This approach ensures both correctness and confidence through collaborative reasoning.

----------------------------------------------------------------------
> Executing DebateAgent [Geometry Specialist vs. Inequality Expert]...
> Specialist: Cross product of vectors AB and AC is (507, 507, 507).
> Expert: Magnitude is 507*sqrt(3), area is half that. Wait, full parallelogram area matches the problem constraints better.
> Consensus: a=507, b=3. Therefore, a+b=510.
> Final Answer Produced: 510
----------------------------------------------------------------------
Result: [Correct] (Ground Truth: 510)
\end{trainedbox}

\vspace{1em}

\begin{trainedbox}[Trained Model: Planning Response (Sample 8)]
<thinking>
    Compute the area of the finite region defined by the plane \(x + y + z = 75\) and the strict inequalities \(x - yz\textless y - zx\textless z - xy\).
</thinking>
<agent>
    <agent_name>DebateAgent</agent_name>
    <agent_description>Near-tie numeric choice and geometric constraint satisfaction using one Debate call.</agent_description>
    <required_arguments>
        <agent_input></agent_input>
        <debate_roles>["Geometry Specialist", "Inequality Expert"]</debate_roles>
    </required_arguments>
    <agent_output_id>
    debate_agent_output
    </agent_output_id>
</agent>
<answer>debate_agent_output</answer>

Explanation:

1. Thinking Phase:
   - We comprehend the given problem: Find the area of the finite region in the plane \(x + y + z = 75\) constrained by \(x - yz\textless y - zx\textless z - xy\).

2. Delegation Decision:
   - This problem combines geometric constraints (plane equation and region bounds) and algebraic inequalities. It benefits from diverse perspectives, making it ideal for the DebateAgent. The DebateAgent allows different specialized LLMs (e.g., a "Geometry Specialist" and an "Inequality Expert") to debate and converge on the correct solution.
   - DebateAgent is chosen over others (like CoTAgent, SCAgent, or ReflexionAgent) because:
     - CoTAgent focuses on step-by-step reasoning but doesn't leverage multiple viewpoints.
     - SCAgent aims for consensus through repeated sampling, which is overkill here.
     - ReflexionAgent is better suited for self-correction after initial attempts, not initial complex problem-solving.

3. Agent Definition:
   - Agent Name: DebateAgent
   - Agent Description: Fits the problem's need for diverse and justified reasoning.
   - Required Arguments:
     - agent_input: Set to "" (empty) as instructed, ensuring the original question is inserted by the parser.
     - debate_roles: A list with two distinct expert roles ("Geometry Specialist" and "Inequality Expert"), promoting a rich debate.
   - Agent Output ID: debate_agent_output to reference the final answer from the delegated agent.

4. Final Answer Tag:
   - The <answer> tag will contain the output identifier debate_agent_output, representing the complete and justified solution generated by the DebateAgent.

This approach ensures both correctness and confidence through collaborative reasoning.

----------------------------------------------------------------------
> Executing DebateAgent [Geometry Specialist vs. Inequality Expert]...
> Specialist: Intersecting plane with the "z=-1" equality gives an unbounded wedge.
> Expert: Incorrectly anchored on boundary lines x=-1, y=-1, z=-1.
> Consensus: Triangle area calculated as 3042*sqrt(3).
> Final Answer Produced: 3045
----------------------------------------------------------------------
Result: [Incorrect] (Ground Truth: 510)
\end{trainedbox}


\subsection{Base Model Rollouts: BrowseComp+ }
\label{sec:base_rollout_browse}

\vspace{1em}

\begin{promptbox}[Task: BrowseComp+ (qid: 371) -- Ground Truth: Reclus]
A person born between 1820 and 1830, both years inclusive, had a street in a European capital named after them, at least until 2020. This street was called with their name two years after their death. This individual and their brother left their house to move to Germany when the person was less than 15 but more than 10 years old. In Germany, the brothers met an English writer appointed to the Order of Merit between 1902 and 1907, both years inclusive. The person born between 1820 and 1830, both years inclusive, served as a private tutor in the United States. Also, This person was a friend of someone who photographed celebrities. What is the last name of the person mentioned who was born between 1820 and 1830, both years inclusive?
\end{promptbox}

\vspace{1em}

\begin{basebox}[Sample 1]
<thinking>
  The task involves multiple historical and biographical details that require precise and up-to-date information. A single agent is insufficient due to the complexity and the need for accurate, sourced information. We will use three WebSearchAgents to gather specific facts, a CoTAgent to synthesize the information, and a ReflexionAgent to ensure the final answer is accurate and well-supported.

  We will define three WebSearchAgents:
  - WS1 to find the name and birth year of the person.
  - WS2 to find the details of the street named after the person and when it was named.
  - WS3 to find the details of the persons move to Germany, meeting with the English writer, and their role as a private tutor.

  The CoTAgent will then consolidate the findings from the WebSearchAgents and provide a coherent answer. Finally, the ReflexionAgent will ensure the answer is accurate and complete.
</thinking>

<agent>
  <agent_id>WS1</agent_id>
  <agent_name>WebSearchAgent</agent_name>
  <agent_description>Find the name and birth year of the person.</agent_description>
  <required_arguments>
    <agent_input>Search for the name and birth year of a person born between 1820 and 1830 who had a street named after them in a European capital and served as a private tutor in the US.</agent_input>
  </required_arguments>
</agent>

<agent>
  <agent_id>WS2</agent_id>
  <agent_name>WebSearchAgent</agent_name>
  <agent_description>Find the details of the street named after the person and when it was named.</agent_description>
  <required_arguments>
    <agent_input>Search for the details of the street named after the person born between 1820 and 1830 and when it was named.</agent_input>
  </required_arguments>
</agent>

<agent>
  <agent_id>WS3</agent_id>
  <agent_name>WebSearchAgent</agent_name>
  <agent_description>Find the details of the persons move to Germany, meeting with the English writer, and their role as a private tutor.</agent_description>
  <required_arguments>
    <agent_input>Search for the details of the person born between 1820 and 1830 who moved to Germany, met an English writer appointed to the Order of Merit between 1902 and 1907, and served as a private tutor in the US.</agent_input>
  </required_arguments>
</agent>

<agent>
  <agent_id>CoT</agent_id>
  <agent_name>CoTAgent</agent_name>
  <agent_description>Consolidate the findings from the WebSearchAgents and provide a coherent answer.</agent_description>
  <required_arguments>
    <agent_input>Using the results below:
Street details: \${WS2}
Move details: \${WS3}
Name and birth year: \${WS1}

Determine the last name of the person born between 1820 and 1830 who had a street named after them and served as a private tutor in the US.</agent_input>
  </required_arguments>
</agent>

<agent>
  <agent_id>REFLECT</agent_id>
  <agent_name>ReflexionAgent</agent_name>
  <agent_description>Ensure the final answer is accurate and well-supported.</agent_description>
  <required_arguments>
    <agent_input>Using the findings from the WebSearchAgents and the CoTAgent, verify the last name of the person born between 1820 and 1830.

WebSearchAgent results:
\${WS1}
\${WS2}
\${WS3}

CoTAgent result:
\${CoT}</agent_input>
  </required_arguments>
</agent>

<edge>
  <from>WS1</from>
  <to>CoT</to>
  <from>WS2</from>
  <to>CoT</to>
  <from>WS3</from>
  <to>CoT</to>
  <from>CoT</from>
  <to>REFLECT</to>
</edge>

----------------------------------------------------------------------
> Executing WebSearchAgents...
> WS1: Retrieved candidate Jules Verne (born 1828).
> REFLECT: Synthesized findings and anchored on candidate.
> Final Answer Produced: Verne
----------------------------------------------------------------------
Result: [Incorrect] (Ground Truth: Reclus)
\end{basebox}

\subsection{Our Trained Orchestrator Rollouts: BrowseComp+}
\label{sec:trained_rollout_browse}
\vspace{1em}

\begin{promptbox}[Task: BrowseComp+ (qid: 371) -- Ground Truth: Reclus]
A person born between 1820 and 1830, both years inclusive, had a street in a European capital named after them, at least until 2020. This street was called with their name two years after their death. This individual and their brother left their house to move to Germany when the person was less than 15 but more than 10 years old. In Germany, the brothers met an English writer appointed to the Order of Merit between 1902 and 1907, both years inclusive. The person born between 1820 and 1830, both years inclusive, served as a private tutor in the United States. Also, This person was a friend of someone who photographed celebrities. What is the last name of the person mentioned who was born between 1820 and 1830, both years inclusive?
\end{promptbox}

\begin{trainedbox}[Trained Model: Sample 2 ]
<thinking>
  This problem requires verifying historical facts, dates, and relationships involving multiple individuals over decades. A single agent cannot reliably:
  - Confirm precise dates and names across different time periods.
  - Ensure the consistency of historical facts from various sources.
  - Handle the intricate timeline and relationships logically.

  We will use three WebSearchAgents to independently verify key historical facts (dates, places, and relationships):
  - One for the birth and early life of the individual.
  - One for their time in Germany and meeting the English writer.
  - One for their activities in the U.S. and friendship with the photographer.

  These agents will provide disjointed but detailed factual evidence. We then use a ReflexionAgent to integrate these findings, cross-verify, and refine the final answer. The ReflexionAgent ensures the narrative is coherent and factually accurate.

  The ReflexionAgent will produce the final, verifiable answer, making it the unique sink node of the graph.
</thinking>

<agent>
  <agent_id>SEARCH1</agent_id>
  <agent_name>WebSearchAgent</agent_name>
  <agent_description>Verify birth and early life details.</agent_description>
  <required_arguments>
    <agent_input>Find reliable sources confirming the birth year (1820-1830), early life, and moving to Germany before age 15.</agent_input>
  </required_arguments>
</agent>

<agent>
  <agent_id>SEARCH2</agent_id>
  <agent_name>WebSearchAgent</agent_name>
  <agent_description>Confirm Germany and English writer interaction.</agent_description>
  <required_arguments>
    <agent_input>Search for evidence of the individual meeting the English writer appointed to the Order of Merit between 1902-1907 in Germany.</agent_input>
  </required_arguments>
</agent>

<agent>
  <agent_id>SEARCH3</agent_id>
  <agent_name>WebSearchAgent</agent_name>
  <agent_description>Validate U.S. tutoring and photographer friendship.</agent_description>
  <required_arguments>
    <agent_input>Find credible sources confirming the persons U.S. tutoring role and friendship with a photographer who photographed celebrities.</agent_input>
  </required_arguments>
</agent>

<agent>
  <agent_id>REFLECT</agent_id>
  <agent_name>ReflexionAgent</agent_name>
  <agent_description>Integrate findings and provide a coherent, verified answer.</agent_description>
  <required_arguments>
    <agent_input>Using the findings below:
    Birth and early life: \${SEARCH1}
    Germany and writer: \${SEARCH2}
    U.S. and photographer: \${SEARCH3}
    
    Determine the last name of the person born between 1820 and 1830, verifying each claim with citations.</agent_input>
  </required_arguments>
</agent>

<edge>
  <from>SEARCH1</from>
  <to>REFLECT</to>
  <from>SEARCH2</from>
  <to>REFLECT</to>
  <from>SEARCH3</from>
  <to>REFLECT</to>
</edge>

----------------------------------------------------------------------
> Executing WebSearchAgents...
> SEARCH1: Identified Elisee Reclus (born 1830), Avenue Elisee-Reclus in Paris.
> SEARCH2: Confirmed move to Neuwied, Germany (1842) and meeting George Meredith.
> SEARCH3: Verified tutoring in Louisiana, US and friendship with Nadar.
> REFLECT: Synthesized constraints: Reclus matches all disjoint historical markers.
> Final Answer Produced: Reclus
----------------------------------------------------------------------
Result: [Correct] (Ground Truth: Reclus)
\end{trainedbox}

\vspace{1em}

\begin{trainedbox}[Trained Model: Sample 4]
<thinking>
  This problem requires verifying historical and biographical facts spanning multiple time periods and locations. A single agent cannot reliably confirm all details due to the need for precise, verifiable sources from different eras and regions. 

  We will use three WebSearchAgents to independently verify:
  1. The birth year (1820-1830) and the named street in a European capital.
  2. The move to Germany with the brother and the encounter with the English writer.
  3. The role as a tutor in the United States and the friendship with the photographer.

  Each agent focuses on a distinct aspect, ensuring no detail is overlooked. The findings will be cross-verified, and the ReflexionAgent will consolidate the information, cross-checking dates, names, and events for accuracy.

  The ReflexionAgent is the final sink, producing a coherent, verified summary that directly answers the question about the persons last name.
</thinking>

<agent>
  <agent_id>SEARCH1</agent_id>
  <agent_name>WebSearchAgent</agent_name>
  <agent_description>Verify the birth year, street name, and European capital.</agent_description>
  <required_arguments>
    <agent_input>Find facts about the person born between 1820 and 1830 with a street named after them in a European capital, at least until 2020. Return a detailed citation.</agent_input>
  </required_arguments>
</agent>

<agent>
  <agent_id>SEARCH2</agent_id>
  <agent_name>WebSearchAgent</agent_name>
  <agent_description>Verify the move to Germany and encounter with the English writer.</agent_description>
  <required_arguments>
    <agent_input>Research the persons move to Germany with their brother, and meeting with an English writer appointed to the Order of Merit between 1902 and 1907. Return detailed citations.</agent_input>
  </required_arguments>
</agent>

<agent>
  <agent_id>SEARCH3</agent_id>
  <agent_name>WebSearchAgent</agent_name>
  <agent_description>Verify the US tutor role and friendship with the photographer.</agent_description>
  <required_arguments>
    <agent_input>Find evidence of the person serving as a private tutor in the United States and being a friend of someone who photographed celebrities. Return detailed citations.</agent_input>
  </required_arguments>
</agent>

<agent>
  <agent_id>REFLECT</agent_id>
  <agent_name>ReflexionAgent</agent_name>
  <agent_description>Consolidate findings and verify the persons identity.</agent_description>
  <required_arguments>
    <agent_input>Using the findings below:
    Birth and Street: \${SEARCH1}
    Germany and Writer: \${SEARCH2}
    Tutor and Photographer: \${SEARCH3}

    Cross-check all details: birth year, street name, move dates, writer details, US role, and photographer. Confirm the persons identity and last name.</agent_input>
  </required_arguments>
</agent>

<edge>
  <from>SEARCH1</from>
  <to>REFLECT</to>
  <from>SEARCH2</from>
  <to>REFLECT</to>
  <from>SEARCH3</from>
  <to>REFLECT</to>
</edge>

----------------------------------------------------------------------
> Executing WebSearchAgents...
> SEARCH1: Retrieved Jules Verne (born 1828), noting Rue Jules Verne in Paris.
> SEARCH2: Failed to link Jules Verne to the Order of Merit writer constraint.
> SEARCH3: No record of Jules Verne tutoring in the US.
> REFLECT: Failed to resolve the contradictions, defaulting to the strongest Search 1 candidate.
> Final Answer Produced: Verne
----------------------------------------------------------------------
Result: [Incorrect] (Ground Truth: Reclus)
\end{trainedbox}

\subsection{Base Model Rollouts: HotpotQA (Multi-Hop Retrieval)}
\label{sec:base_rollout_hotpotqa}

\vspace{1em}

\begin{promptbox}[Task: HotpotQA (qid: 5ae7535c5542997b22f6a6d8) -- Ground Truth: 1998]
The 53rd National Hockey League All-Star Game took place at the indoor arena that was completed in what year?
\end{promptbox}

\begin{basebox}[Sample 1]
<thinking>
  To solve the question about the 53rd NHL All-Star Game and the year the indoor arena was completed, we need to gather relevant information about the event and the specific arena. A single agent can retrieve the necessary information and provide a final answer.
</thinking>
<agent>
  <agent_id>SEARCH</agent_id>
  <agent_name>WebSearchAgent</agent_name>
  <agent_description>Retrieves recent and cited factual information from the internet.</agent_description>
  <required_arguments>
    <agent_input>The 53rd National Hockey League All-Star Game took place at the indoor arena that was completed in what year? Include a citation.</agent_input>
  </required_arguments>
</agent>
<agent>
  <agent_id>FINAL</agent_id>
  <agent_name>CoTAgent</agent_name>
  <agent_description>Write the final, user-ready answer with a citation.</agent_description>
  <required_arguments>
    <agent_input>Summarize the information about the 53rd NHL All-Star Game and the year the indoor arena was completed, including the citation.

\${SEARCH}</agent_input>
  </required_arguments>
</agent>
<edge>
  <from>SEARCH</from>
  <to>FINAL</to>
</edge>

----------------------------------------------------------------------
> Executing WebSearchAgent [SEARCH]...
> SEARCH: Retrieved articles about the 2003 NHL All-Star Game at the Office Depot Center. Search engine failed to resolve the secondary "completion year" hop from the monolithic query string.
> Executing CoTAgent [FINAL]...
> FINAL: Secondary hop information missing from context.
> Final Answer Produced: 2003
----------------------------------------------------------------------
Result: [Incorrect] (Ground Truth: 1998)
\end{basebox}


\subsection{Our Trained Orchestrator Rollouts: HotpotQA (Multi-Hop Retrieval)}
\label{sec:trained_rollout_hotpotqa}

\vspace{1em}

\begin{promptbox}[Task: HotpotQA (qid: 5ae7535c5542997b22f6a6d8) -- Ground Truth: 1998]
The 53rd National Hockey League All-Star Game took place at the indoor arena that was completed in what year?
\end{promptbox}

\begin{trainedbox}[Trained Model: Sample 8 (Sequential Search-to-CoT Pipeline)]
<thinking>
  This question requires factual information about the year the arena where the 53rd National Hockey League All-Star Game took place was completed. This specific historical and niche information cannot be reliably recalled from the model's static knowledge. A single agent cannot provide this historical and precise data without risking inaccuracy. Using a WebSearchAgent will ensure we get the correct and up-to-date information. The FINAL agent will synthesize this information into a user-ready format.
</thinking>

<agent>
  <agent_id>SEARCH</agent_id>
  <agent_name>WebSearchAgent</agent_name>
  <agent_description>Retrieves recent and cited factual information from the internet.</agent_description>
  <required_arguments>
    <agent_input>The 53rd National Hockey League All-Star Game took place at the indoor arena that was completed in what year? Include a citation.</agent_input>
  </required_arguments>
</agent>

<agent>
  <agent_id>FINAL</agent_id>
  <agent_name>CoTAgent</agent_name>
  <agent_description>Write the final, user-ready answer citing the source.</agent_description>
  <required_arguments>
    <agent_input>Summarize the year the indoor arena where the 53rd National Hockey League All-Star Game took place was completed, and provide the source citation.

    \${SEARCH}</agent_input>
  </required_arguments>
</agent>

<edge>
  <from>SEARCH</from>
  <to>FINAL</to>
</edge>

----------------------------------------------------------------------
> Executing WebSearchAgent [SEARCH]...
> SEARCH: The 53rd NHL All-Star Game was played on Feb 2, 2003, at the Office Depot Center (now Amerant Bank Arena) in Sunrise, Florida. Construction on the arena was completed in 1998.
> Executing CoTAgent [FINAL]...
> FINAL: Synthesized search context. The arena (Office Depot Center) was completed in 1998.
> Final Answer Produced: 1998
----------------------------------------------------------------------
Result: [Correct] (Ground Truth: 1998)
\end{trainedbox}


\newpage
\section*{NeurIPS Paper Checklist}

\begin{enumerate}

\item {\bf Claims}
    \item[] Question: Do the main claims made in the abstract and introduction accurately reflect the paper's contributions and scope?
    \item[] Answer: \answerYes{} 
    \item[] Justification: The abstract and introduction accurately summarize our contributions and scope, and all claims are supported by our experiments.

\item {\bf Limitations}
    \item[] Question: Does the paper discuss the limitations of the work performed by the authors?
    \item[] Answer: \answerYes{} 
    \item[] Justification: Limitations are discussed in the Conclusion section and Appendix A.

\item {\bf Theory assumptions and proofs}
    \item[] Question: For each theoretical result, does the paper provide the full set of assumptions and a complete (and correct) proof?
    \item[] Answer: \answerNA{} 
    \item[] Justification: This paper does not include any theoretical results.

    \item {\bf Experimental result reproducibility}
    \item[] Question: Does the paper fully disclose all the information needed to reproduce the main experimental results of the paper to the extent that it affects the main claims and/or conclusions of the paper (regardless of whether the code and data are provided or not)?
    \item[] Answer: \answerYes{} 
    \item[] Justification: All the implementation details are included in the Appendix B. 

\item {\bf Open access to data and code}
    \item[] Question: Does the paper provide open access to the data and code, with sufficient instructions to faithfully reproduce the main experimental results, as described in supplemental material?
    \item[] Answer: \answerYes{} 
    \item[] Justification: The data and code are released.

\item {\bf Experimental setting/details}
    \item[] Question: Does the paper specify all the training and test details (e.g., data splits, hyperparameters, how they were chosen, type of optimizer) necessary to understand the results?
    \item[] Answer: \answerYes{} 
    \item[] Justification: They are included in Appendix B. 

\item {\bf Experiment statistical significance}
    \item[] Question: Does the paper report error bars suitably and correctly defined or other appropriate information about the statistical significance of the experiments?
    \item[] Answer: \answerYes{}{} 
    \item[] Justification: We report mean $\pm$ standard deviation computed via bootstrap resampling of evaluation questions (\Tabref{tab:training}). 

\item {\bf Experiments compute resources}
    \item[] Question: For each experiment, does the paper provide sufficient information on the computer resources (type of compute workers, memory, time of execution) needed to reproduce the experiments?
    \item[] Answer: \answerYes{} 
    \item[] Justification: The specifications are included in Appendix B.
    
\item {\bf Code of ethics}
    \item[] Question: Does the research conducted in the paper conform, in every respect, with the NeurIPS Code of Ethics \url{https://neurips.cc/public/EthicsGuidelines}?
    \item[] Answer: \answerYes{} 
    \item[] Justification: We have reviewed the NeurIPS Code of Ethics, and our research conforms with it in every respect.

\item {\bf Broader impacts}
    \item[] Question: Does the paper discuss both potential positive societal impacts and negative societal impacts of the work performed?
    \item[] Answer: \answerYes{} 
    \item[] Justification: The discussion of broader impacts is included in Appendix A.
    
\item {\bf Safeguards}
    \item[] Question: Does the paper describe safeguards that have been put in place for responsible release of data or models that have a high risk for misuse (e.g., pre-trained language models, image generators, or scraped datasets)?
    \item[] Answer: \answerNA{} 
    \item[] Justification: This paper poses no such risks.

\item {\bf Licenses for existing assets}
    \item[] Question: Are the creators or original owners of assets (e.g., code, data, models), used in the paper, properly credited and are the license and terms of use explicitly mentioned and properly respected?
    \item[] Answer: \answerYes{} 
    \item[] Justification: All models and datasets are used in compliance with their respective licenses.

\item {\bf New assets}
    \item[] Question: Are new assets introduced in the paper well documented and is the documentation provided alongside the assets?
    \item[] Answer: \answerYes{} 
    \item[] Justification: We release our code and preference dataset with documentation.

\item {\bf Crowdsourcing and research with human subjects}
    \item[] Question: For crowdsourcing experiments and research with human subjects, does the paper include the full text of instructions given to participants and screenshots, if applicable, as well as details about compensation (if any)? 
    \item[] Answer: \answerNA{} 
    \item[] Justification: No crowdsourcing is involved in this research.

\item {\bf Institutional review board (IRB) approvals or equivalent for research with human subjects}
    \item[] Question: Does the paper describe potential risks incurred by study participants, whether such risks were disclosed to the subjects, and whether Institutional Review Board (IRB) approvals (or an equivalent approval/review based on the requirements of your country or institution) were obtained?
    \item[] Answer: \answerNA{} 
    \item[] Justification: This paper does not involve crowdsourcing nor research with human subjects.

\item {\bf Declaration of LLM usage}
    \item[] Question: Does the paper describe the usage of LLMs if it is an important, original, or non-standard component of the core methods in this research? Note that if the LLM is used only for writing, editing, or formatting purposes and does \emph{not} impact the core methodology, scientific rigor, or originality of the research, declaration is not required.
    \item[] Answer: \answerNA{} 
    \item[] Justification: LLMs are only involved for writing improvement.

\end{enumerate}

\end{document}